\documentclass[10pt,twocolumn,letterpaper]{article}

\usepackage{iccv}

\makeatletter
\@namedef{ver@everyshi.sty}{}
\makeatother
\usepackage{tikz}

\usepackage{times}
\usepackage{epsfig}
\usepackage{graphicx}
\usepackage{amsmath}
\usepackage{amssymb}

\usepackage{booktabs}
\usepackage{array}
\newcolumntype{L}[1]{>{\raggedright\let\newline\\\arraybackslash\hspace{0pt}}m{#1}}
\newcolumntype{C}[1]{>{\centering\let\newline\\\arraybackslash\hspace{0pt}}m{#1}}
\newcolumntype{R}[1]{>{\raggedleft\let\newline\\\arraybackslash\hspace{0pt}}m{#1}}
\usepackage{tablefootnote}
\usepackage[font=small]{caption}
\usepackage{subcaption}
\usepackage{multirow}
\usepackage{enumitem}
\usepackage{float}
\usepackage{xcolor}
\usepackage{algorithm2e}
\RestyleAlgo{ruled}
\SetKwInput{kwInput}{Input}
\usepackage{pgfplots}
\pgfplotsset{compat=1.9}

\definecolor{bblue}{HTML}{5B9BD5}
\definecolor{ggreen}{HTML}{70AD47}
\definecolor{ggray}{HTML}{A5A5A5}

\usepackage[pagebackref=true,breaklinks=true,letterpaper=true,colorlinks,bookmarks=false]{hyperref}
\usepackage[toc,page]{appendix}

\iccvfinalcopy 

\def\iccvPaperID{****} 

\makeatletter
\let\ps@empty\ps@plain
\makeatother

\begin{document}

\title{Versatile Diffusion: Text, Images and Variations All in One Diffusion Model}

\author{
    Xingqian Xu\textsuperscript{1},
    Zhangyang Wang\textsuperscript{2,3},
    Eric Zhang\textsuperscript{1},
    Kai Wang\textsuperscript{1},
    Humphrey Shi\textsuperscript{1,3} \\
{\small \textsuperscript{1}SHI Labs @ UIUC, Georgia Tech \& U of Oregon, \textsuperscript{2}UT Austin, \textsuperscript{3}Picsart AI Research (PAIR)}\\
{\small \textbf{\url{https://github.com/SHI-Labs/Versatile-Diffusion}}}
}

\twocolumn[{
\maketitle
\begin{center}
    \captionsetup{type=figure}
    \includegraphics[width=0.99\textwidth]{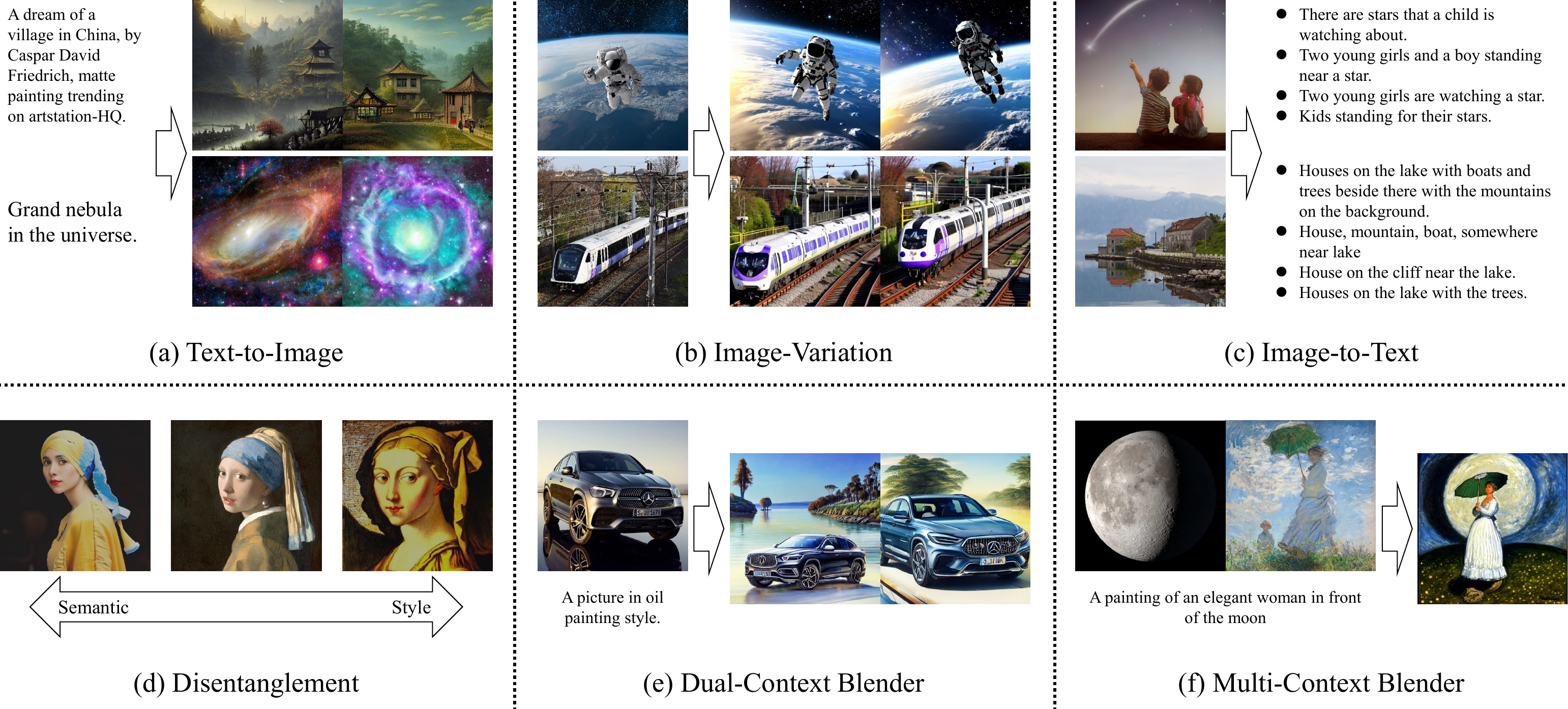}
    \captionof{figure}{Demo results of our Versatile Diffusion (\textbf{VD}) framework on three out of all primary tasks (\ie Figures \textbf{a}, \textbf{b}, and \textbf{c}) and three derived tasks (\ie Figure \textbf{d}, \textbf{e}, and \textbf{f}). As shown in the captions, the three primary tasks are text-to-image, image-variation, and image-to-text. Figure \textbf{d} demonstrates the disentanglement between image semantics and style. Figure \textbf{e} shows the demo of dual-context blender using one image and one text. Figure \textbf{f} shows the demo of the multi-context blender using multiple images and one text.}
    \label{fig:teasor}
    \vspace{0.4cm}
\end{center}
}]
\thispagestyle{empty}

\maketitle
\ificcvfinal\thispagestyle{empty}\fi

\begin{abstract}

Recent advances in diffusion models have set an impressive milestone in many generation tasks, and trending works such as DALL-E2, Imagen, and Stable Diffusion have attracted great interest. Despite the rapid landscape changes, recent new approaches focus on extensions and performance rather than capacity, thus requiring separate models for separate tasks. In this work, we expand the existing single-flow diffusion pipeline into a \textbf{multi-task multimodal} network, dubbed \textbf{Versatile Diffusion} (VD), that handles multiple flows of text-to-image, image-to-text, and variations in one unified model. The pipeline design of VD instantiates a \textbf{unified multi-flow diffusion framework}, consisting of sharable and swappable layer modules that enable the crossmodal generality beyond images and text. Through extensive experiments, we demonstrate that VD successfully achieves the following: a) VD outperforms the baseline approaches and handles all its base tasks with competitive quality; b) VD enables novel extensions such as disentanglement of style and semantics, dual- and multi-context blending, \etc; c) The success of our multi-flow multimodal framework over images and text may inspire further diffusion-based universal AI research. Our code and models are open-sourced at \href{https://github.com/SHI-Labs/Versatile-Diffusion}{https://github.com/SHI-Labs/Versatile-Diffusion}.

\end{abstract}

\vspace{-0.5cm}
\section{Introduction}\label{sec:intro}
Multi-modality is the ``crown jewel" for achieving universal AI. 
With the attributes of deep learning, methods designed for traditional tasks such as classification, detection, segmentation, \etc, have reached near-human level accuracy. On top of them, multimodal research such as ~\cite{mm_zscls0, mm_cls1, mm_cls2, oneformer} primarily focused on discriminative tasks of jointly recognizing, matching, or understanding multimodal data. Nevertheless, research on multimodal generative models remains scarce. Previously, the best-performing generative vision models, generative adversarial networks (GAN)~\cite{progressivegan, biggan, stylegan2} merely focus on specific domains (\ie faces~\cite{stylegan2, pigan, stylenat}, fonts~\cite{shapemgan, textstylebrush}, natural scenes~\cite{singan, infinitygan}, \etc); and on specific tasks (inpainting \cite{lama, comodgan, shgan}, super-resolution \cite{ledig2017photo}, image-to-image translation~\cite{cgan, cyclegan}, \etc). 

The recent success of diffusion models~\cite{ddpm, ddim, dalle2, imagen, ldm} has brought new horizons. Diffusion models are likelihood-based models that gradually restore image contents from Gaussian corruptions. It has proved to be effective in bridging modalities and tasks, for instance, unconditional generation~\cite{ddpm, ddim, dm_beat_gan}, density estimation~\cite{dm_vdm}, super-resolution~\cite{sr3}, and text-to-image generation~\cite{glide, dalle2, imagen, ldm}. The success of diffusion models can be attributed to several aspects. Firstly, their training objectives lead to a more robust training procedure than other approaches like GANs. The iterative refinement inference procedure also expands the model capability at the cost of more running time. Besides, the competitive performance of recent diffusion models such as DALL-E2~\cite{dalle2}, Imagen~\cite{imagen}, and Stable Diffusion~\cite{ldm} benefits from the remarkable data collection such as LAION~\cite{laion}, CC12M~\cite{cc12m}, COYO~\cite{coyo}, etc. 
The disadvantages of earlier diffusion models, such as the data hunger and high inference costs, are gradually alleviated by more efficient structures and schedulers~\cite{ddim, dm_fast0, dm_fast1, dm_fast2, ldm}. Diffusion-based text-to-image methods~\cite{dalle2, imagen, ldm} arguably set new state-of-the-art for multi-modal generative AI. However, those works by far almost exclusively hinge on single-flow diffusion pipelines (illustrated in Section 3); and meanwhile, most of them are trained and evaluated on a single specialized generation task (e.g., text to image) despite being cross-modality. 



\textit{What is the next move forward, then?} We believe in the central role of multimodal, multi-task models in universal AI, and we consider diffusion models to be a promising workhorse to enable so. 
To fulfill our goal, we proposed \textit{Versatile Diffusion} (\textbf{VD}) that comprehensively solves text, images, and variations within one unified generative model. The \textit{key underlying technique} is a novel multi-flow diffusion framework, that generalizes existing single-flow diffusion pipelines to \textit{handle multiple modalities and tasks simultaneously} while effectively sharing information across them. Thanks to the \textit{larger capacity} as well as capturing crossmodal semantics, VD not only performs well on the aforementioned supported tasks but notably derives many new capabilities including semantic-style disentanglement, cross-modal dual context or multi-context generation (blending), leading to remarkable advances of \textit{empirical performance} for multi-modal generative AI. Our main contributions are summarized in the following:
\vspace{-0.1cm}

\begin{itemize}
    \item We introduce \textit{Versatile Diffusion} (\textbf{VD}), a multimodal, multi-task diffusion network that adopts a novel generalized multi-flow pipeline, unlike existing single-flow diffusion models.
    \vspace{-0.2cm}
    \item VD solves multiple modalities and tasks in one unified model, including image generation (text-to-image, image-variation), and text generation (image-to-text, text-variation). Through comprehensive experiments, we show that VD outperforms the baselines via scores and quality. For example, VD's high-quality text-to-image and image-variation results demonstrate that it indeed better captures the context semantics.
    \vspace{-0.2cm}
    \item The unique multi-flow multimodal property of VD enables more novel derivative tasks, that may further facilitate downstream users engaged in this technology, including the semantic-style disentanglement, dual-context and multi-context blending, \etc.
\end{itemize}

\section{Related Works}\label{sec:related}
\textbf{Multi-modalities} are unions of information with different forms, including but not limited to vision, text, audio, \etc~\cite{mm_book0, mm_survey0}. Early deep learning work led by Ngiam \etal ~\cite{mm_dl0} learned a fused representation for audio and video. The similar idea was also adopted across vision and text label~\cite{mm_dl0}, and across vision and language~\cite{mm_dl2}. A part of multimodal approaches focused on zero-shot learning, for instance, DiViSE~\cite{mm_zscls0} targeted mapping images on semantic space from which unseen category labels can be predicted. Socher \etal~\cite{mm_zscls1} trained a recognition model with similar ideas in which images were projected on the space of text corpus. \cite{mm_zscls2} shared the same design as DiViSE but was upgraded for a large and noisy dataset. Another set of works~\cite{mm_cls0, mm_cls1,  mm_cls2,  mm_cls3}, focused on increasing classification accuracy via multimodal training: in which~\cite{mm_cls0} and~\cite{mm_cls1} did a simple concatenation on multimodal embeddings; ~\cite{mm_cls2} proposed a gated unit to control the multimodal information flow in the network; ~\cite{mm_cls3} surveyed FastText~\cite{fasttext} with multiple fusion methods on text classification. Meanwhile, multimodal training was also wide-adopted in detection and segmentation~\cite{rcnn,maskrcnn, oneformer}
in one shot. Another topic, VQA~\cite{mm_vqa0, mm_vqa1}, conducted cross-modal reasoning that transferred visual concepts into linguistic answers. Methods such as~\cite{mm_vqa2, mm_vqa3} extracted visual concepts into neural symbolics, and ~\cite{mm_vqa4, mm_vqa5} learned additional concept structures and hierarchies.

\textbf{Multimodal generative tasks} involve simultaneous representation learning and generation/synthesis~\cite{mm_survey1}, in which representation networks~\cite{ae, vae, gan, vqvae, wavenet, prnet} with contrastive loss~\cite{clip, cl, mm_cl0, mm_cl1, mm_cl2} played an essential role. Specifically, our model VD adopts VAEs~\cite{vae} and CLIP~\cite{clip} as the latent and context encoders, which are two critical modules for the network. VD also shares the common cross-modal concepts such as domain transfer~\cite{cgan, cyclegan} and joint representation learning~\cite{mm_dl1, mm_gm0, mm_gm1}.

\textbf{Diffusion models} (DM)~\cite{dm_early0, ddpm} consolidate large family of methods including VAEs~\cite{vae, vqvae, vqvae2}, Markov chains~\cite{mcm0, dm_early0, mcm1, mcm2}, and score matching models~\cite{scorem0, scorem1}, \etc. Differ from GAN-based\cite{gan, biggan, stylegan2} and flow-based models~\cite{flow0, flow1}, DM minimizes the lower-bounded likelihoods~\cite{ddpm, scorem0} in backward diffusion passes, rather than exact inverse in flow~\cite{flow0} or conduct adversarial training~\cite{gan}. Among the recent works, DDPM~\cite{ddpm} prompted $\epsilon$-prediction that established a connection between diffusion and score matching models via annealed Langevin dynamics sampling~\cite{dm_early1, scorem0}. DDPM also shows promising results on par with GANs in unconditional generation tasks. Another work, DDIM~\cite{ddim}, proposed an implicit generative model that yields deterministic samples from latent variables. Compared with DDPM, DDIM reduces the cost of sampling without losing quality. Regarding efficiency, FastDPM~\cite{dm_fast0} investigated continuous diffusion steps and generalized DDPM and DDIM with faster sampling schedules. Another work, ~\cite{dm_fast1}, replaced the original fixed sampling scheme with a learnable noise estimation that boosted both speed and quality. ~\cite{dm_fast2} introduced a hieratical structure with progressive increasing dimensions that expedite image generations for DM. Regarding quality, ~\cite{dm_beat_gan} compared GANs with DMs with exhaustive experiments and concluded that DMs outperformed GANs on many image generation tasks. Another work, VDM~\cite{dm_vdm}, introduced a family of DM models that reaches state-of-the-art performance on density estimation benchmarks. Diffwave~\cite{dm_diffwave} and WaveGrad~\cite{dm_wavegrad} show that DM also works well on audio. ~\cite{dm_improved_ddpm} improved DDPM with learnable noise scheduling and hybrid objective, achieving even better sampling quality. 
\cite{dm_morecontrol} introduced semantic diffusion guidance to allow image or language-conditioned synthesis with DDPM.

\textbf{Text-to-image generation}, nowadays a joint effort of multimodal and diffusion research, has drawn lots of attention. Among these recent works, GLIDE~\cite{glide} adopted pretrained language models and the cascaded diffusion structure for text-to-image generation. DALL-E2~\cite{dalle2}, a progressive version from DALL-E~\cite{dalle}, utilized CLIP model~\cite{clip} to generate text embedding and adopted the similar hieratical structure that made 256 text-guided images and then upscaled to 1024. Similarly, Imagen~\cite{imagen} explored multiple text encoders~\cite{bert, t5, clip} with conditional diffusion models and explores the trade-offs between content alignment and fidelity via various weight samplers. LDM~\cite{ldm} introduced a novel direction in which the model diffuses on VAE latent spaces instead of pixel spaces. Such design reduced the resource needed during inference time, and its latter version, SD, has proven to be equally effective in text-to-image generation.

\section{Method}\label{sec:method}
In this section, we will first revisit the fundamentals of diffusion models~\cite{dm_early0,ddpm}, including the forward-backward processes and training objectives. We will then highlight the multi-flow multimodal framework of Versatile Diffusion (VD), which is a key contribution that makes VD a unified model of multiple tasks. Finally, we will reveal all details of VD, including the choice of VAEs, context encoders, loss functions, \etc.

\subsection{Diffusion basics}

The forward diffusion process $p(x_T|x_0)$ is a Markov Chain~\cite{ddpm} with $T$ steps that gradually degrade $x_0$ to $x_T$ with random Gaussian noises (Equation~\ref{eq:forward_diffusion}).

\vspace{-0.3cm}
\begin{equation}\begin{gathered}
    q(x_T|x_0) = \prod_{t=1}^T q(x_{t}|x_{t-1}) = \prod_{t=1}^T \mathcal{N}(\sqrt{1-\beta_t}x_{t-1}; \beta_t\mathbf{I})\\
    = \mathcal{N}(\sqrt{\bar{\alpha}_t}x_0; (1-\bar{\alpha}_t\mathbf{I}));\\
    \bar{\alpha}_t =\prod_{t=1}^T\alpha_t; 
    \quad
    \alpha_t = 1-\beta_t
\label{eq:forward_diffusion}
\end{gathered}\end{equation}

Given the forward diffusion process as prior, diffusion models are trained to reverse the process and recover signal $x_0$ back from $x_T$ by removing the added Gaussian noises. This is known as the backward diffusion process, and each step $p_{\theta}(x_{t-1}|x_t)$ is sampled from the Gaussian distribution with network predicted mean $\mu_{\theta}(x_t, t)$ and variance $\Sigma_{\theta}(x_t, t)$, shown as Equation~\ref{eq:backward_diffusion}.

\vspace{-0.1cm}
\begin{equation}\begin{gathered}
    p_{\theta}(x_{t-1}|x_t) = \mathcal{N}(
        \mu_{\theta}(x_t, t),
        \Sigma_{\theta}(x_t, t)
    )
\label{eq:backward_diffusion}
\end{gathered}\end{equation}

The objective function to train a diffusion model is to minimize the variational bound for negative log-likelihood~\cite{ddpm} shown in Equation~\ref{eq:objective}. In practice, many works assume deterministic $\alpha_t$ and $\beta_t$ for step $t$ in Equation~\ref{eq:forward_diffusion}. Given that both forward and backward processes are Gaussian processes, the objective can then be simplified as the variational weighted $l_2$ loss between the ground truth and predicted mean. 

\vspace{-0.3cm}
\begin{equation}\begin{gathered}
    L = \mathbb{E}[-\log p_{\theta}(x_0)] 
    \le \mathbb{E}\left[
        -\log\frac{p_{\theta}(x_{0:T})}{q(x_{1:T}|x_0)}
    \right]
\label{eq:objective}
\end{gathered}\end{equation}

\vspace{-0.1cm}
\subsection{Multi-flow multimodal diffusion framework}\label{sec:framework}

\begin{figure}[t]
    \centering
    \includegraphics[width=0.495\textwidth]{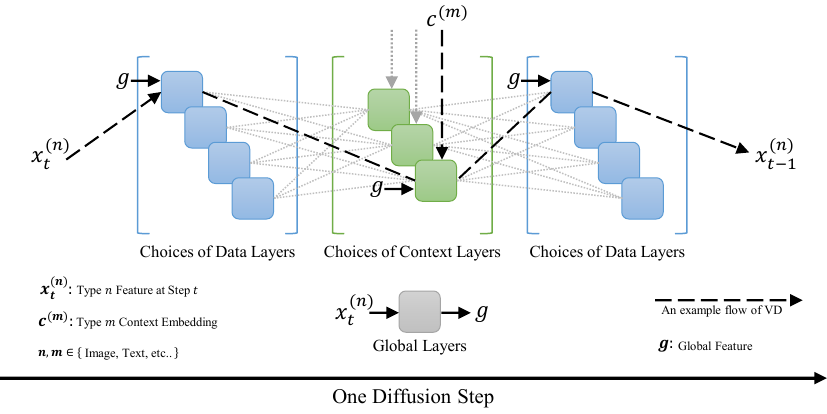}
    \vspace{-0.3cm}
    \caption{Graphic illustration of one diffusion step of VD's multi-flow multimodal diffusion framework, in which multiple choices of data layers (blue), context layers (green), and fixed global layers (gray) are involved. The black dash line shows one flow of the model that handles one crossmodal task (\ie text-to-image), in which the top data blocks, the bottom context blocks, and the shared global layers are activated.  Other data and context blocks stay silent but will be activated when performing other tasks.}
    \vspace{-0.4cm}
    \label{fig:framework}
\end{figure}

The core part of Versatile Diffusion (VD) is the multi-flow multimodal diffusion framework capable of generating various forms of outputs (\eg image, text, 3D, \etc) conditioned on various crossmodal contexts (\eg image, text, audio \etc). A formal definition of a single flow in VD is to synthesize features of modality $n$ using contexts of modality $m$. One may notice that the well-explored text-to-image task~\cite{mask-a-scene, dalle2, imagen, ldm}, \ie synthesizing images based on text prompts, matches the definition of a single flow in VD. But the scope of VD goes beyond one single task; particularly in this work, VD is set up to fulfill numerous tasks: text-to-image, image-to-text, and variations, and may further extend to cover more modalities such as 3D, audio, music, \etc.

Speaking with details, VD handles groups of crossmodal tasks due to its multi-flow framework, in which layers can be activated or muted based on the modalities of the input contexts and output results. As shown in Figure~\ref{fig:framework}, we categorize all diffuser layers into three groups: global layers, data layers, and context layers. The global layers are flow-independent layers that will always be activated. Data layers are output-dependent layers that will be activated when the network generates the corresponding output type. Lastly, context layers are context-dependent layers that will be activated when the corresponding context type is input. Using SD~\cite{ldm} as a reference, the global layers are time-embedding layers; the data layers are residual blocks; and the context layers are cross-attentions. One flow of VD routes the feed-forward pass through the shared global layers and the chosen data and context layers, while other irrelevant layers will stay silent (see Figure~\ref{fig:framework}). Use text-to-image as an example. The $t$-step intermediate result $x_t$ will be fed to image data blocks and text context blocks to generate the next step result $x_{t-1}$. Similarly, if our goal is to perform image-variation, we need to use image data blocks and image context blocks. 

One may notice that such a multi-flow multimodal framework highly promotes parameter sharing. In this work, our default VD setting is a four-flow model. In order to replicate such four-flow VD, one would require a total of four diffusion models (\ie four times the size of an SD~\cite{ldm}), while VD reduces the number of parameters by half via its shared layers in the framework. A more generalized version of VD handles $N\times M$ crossmodal tasks with $N$ types of output and $M$ types of context. The size of the model would then become $\mathcal{O}\left(\text{max}(N, M)\right)$, which is significantly smaller than a vanilla model ensembling that requires an accumulated size of $\mathcal{O}(N\times M)$.

\begin{figure*}[t]
    \centering
    \includegraphics[width=0.96\textwidth]{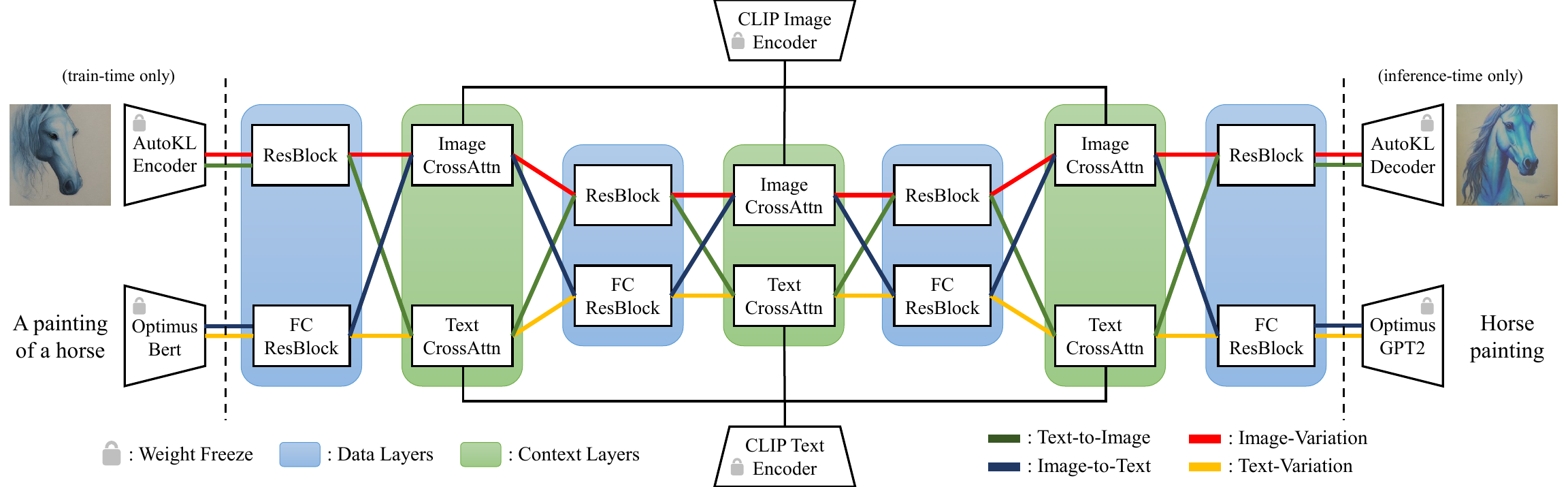}
    \caption{
        The overall structure of four-flow Versatile Diffusion (VD). Each color line depicts a single flow of VD that represents one supported task (\ie green line for text-to-image). The VAE encoders at the far left are only used in training and are replaced with Gaussian noise inputs during inference. Oppositely, the VAE decoders at the far right are only used in inference for output generation, not train-time loss computation. For simplicity, we hide global layers in this figure. Better viewed in color.}
    \vspace{-0.1cm}
    \label{fig:vd}
\end{figure*}

\vspace{0.1cm}

\subsection{Versatile Diffusion}

\textbf{Tasks}: As mentioned earlier, Versatile Diffusion (VD) is a unified diffusion model for text-to-image, image-to-text, and variations. Text-to-image and image-to-text are two well-known tasks in which the former generates images from text prompts, and the latter generates image captioning. Image-variation (IV) is a fairly new task in which users generate new images that are semantically similar to the reference images. IV differs from SD's image-to-image (I2I)~\cite{ldm} by two points a) IV diffuses from pure noise while I2I diffuses from images half-mixed with noise; b) IV maintains high-level semantics but relaxes the low-level structures, while I2I only replicates low-level structures and has no guarantee on high-level semantics. Lastly, VD can also generate variations in text due to its multi-flow nature, whose goal is to generate similar expressions from reference text.

\vspace{0.1cm}

\begin{figure}[t]
    \centering
    \includegraphics[width=0.48\textwidth]{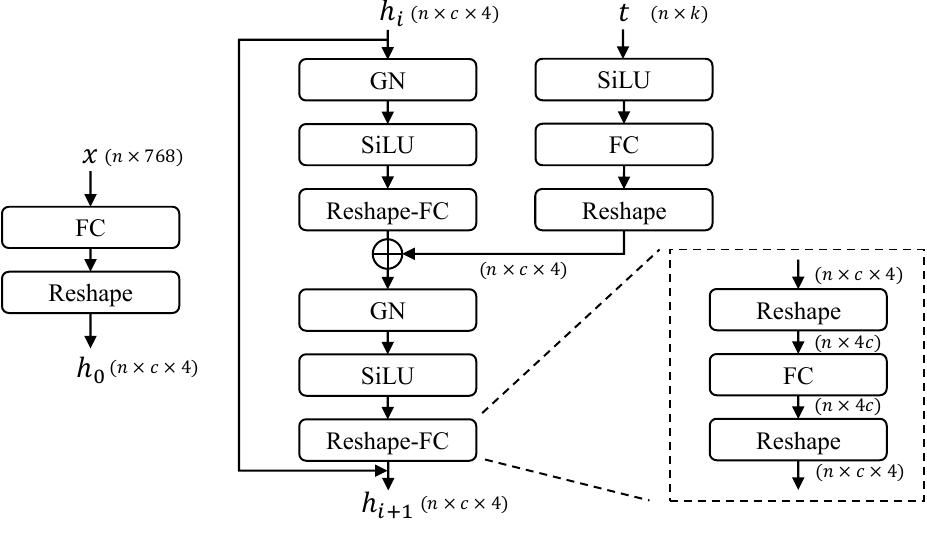}
    \vspace{-0.6cm}
    \caption{FCResBlock contains two sets of fully connected layers (FC), group normalizations (GN)~\cite{groupnorm}, and sigmoid linear units (SiLU)~\cite{silu}. $x$ is the input text latent code, $t$ is the input time embedding, and $h_i$ are the intermediate features.}
    \label{fig:fcresblock}
\end{figure}

\textbf{Network}: The full model of VD includes three components: a) A diffuser that follows our multi-flow multimodal framework described in Sec~\ref{sec:framework}; b) VAEs that convert data samples to latent representations; c) Context encoders that encode contexts into embeddings. The overall network diagram is also shown in Figure~\ref{fig:vd}.
\textbf{Diffuser}: We use the well-adopted UNet~\cite{unet} with cross attentions~\cite{attn} as the main structure of our diffuser network. Part of the UNet follows SD~\cite{ldm}, where we adopt residual blocks~\cite{resnet} as image data layers and cross-attention as text and image context layers. For text data layers, we propose the fully connected residual blocks (FCResBlock) that expand 768-dimensional text latent vectors into a 320-by-4 hidden feature and follow a similar residual block paradigm with GroupNorms~\cite{groupnorm}, SiLU~\cite{silu}, and skip connections (see Figure~\ref{fig:fcresblock}). 
\textbf{VAE}: We adopt the same Autoencoder-KL~\cite{ldm} like SD as our image VAE. Parallelly, we adopt Optimus~\cite{optimus} as our text VAE. Optimus consists of a Bert~\cite{bert} text encoder and a GPT2~\cite{gpt2} text decoder, by which it can bidirectionally transform sentences into 768-dimensional normally-distributed latent vectors. 
\textbf{Context encoder}: We use both CLIP~\cite{clip} text and image encoders as VD's context encoders. Unlike SD, which uses raw text embeddings as context inputs, we use normalized and projected embeddings that minimize the CLIP text-image contrastive loss. In our experiments, we noticed that closer embedding spaces between contexts (\ie image and text) help converge fast and perform better. 

\begin{algorithm}
\caption{Backpropagation of VD}
    $X=\{x^{(1)} \ldots x^{(N)}\}$; \tcp{N types data}
    $C=\{c^{(1)} \ldots c^{(M)}\}$; \tcp{M types context}
    $L_{\theta}(x^{(\cdot)},c^{(\cdot)})$; \tcp{Loss with params $\theta$}
    $\delta_\theta = 0$; \tcp{Param gradients}
    \For{$x^{(i)} \in X$}{
        \For{$c^{(j)} \in C$}{
            $\delta'_\theta = \nabla_{\theta}L_\theta(x^{(i)}, c^{(j)})$; \tcp{One flow}
            $\delta_\theta = \delta_\theta + \delta'_\theta$\;
        }
    }
    Update network with $\delta_\theta$\;
\label{alg:loss}
\end{algorithm}
\setlength{\textfloatsep}{0.4cm}

\textbf{Loss}: Training VD is surprisingly simple. For each of the flows, we compute the variational weighted $l_2$ losses described in Equation~\ref{eq:objective} and do regular backpropagation (see Algorithm~\ref{alg:loss}). Model weights will be updated when the gradients in all flows are accumulated. Besides, when updating the weights, we manually set gradient scales for parameters in data and context layers to better adapt our multi-flow model settings. More information can be found in the Experiments session.

\section{Experiments}\label{sec:experiments}
In this session, we will describe VD's data and settings, show the performance of VD on primary tasks, and introduce several derived applications empowered by the multi-flow multimodal property of VD.

\begin{figure*}[t]
    \centering
    \begin{subfigure}[b]{0.90\textwidth}
        \centering
        \includegraphics[width=\textwidth]{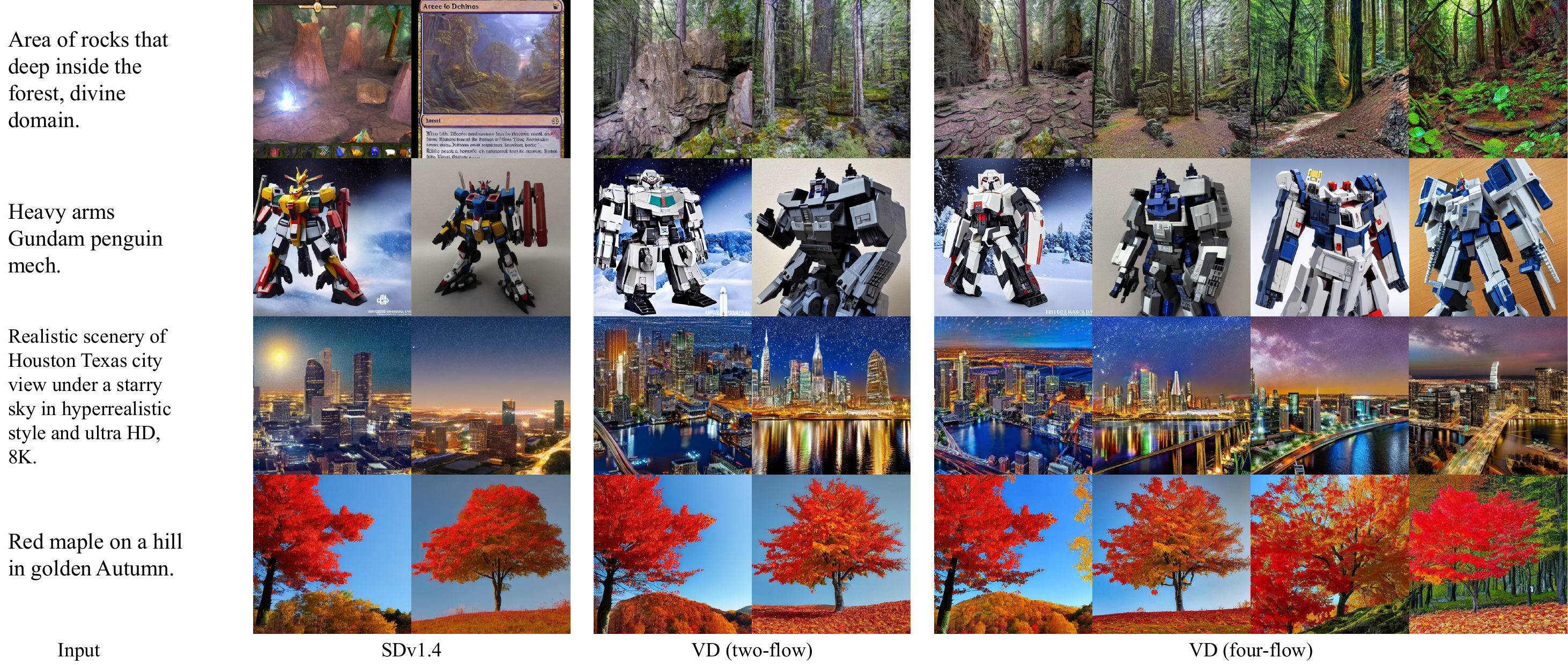}
        \caption{Text-to-Image performance.}
        \vspace{0.2cm}
        \label{fig:qcompare_t2i}
    \end{subfigure}
    
    \begin{subfigure}[b]{0.90\textwidth}
        \centering
        \includegraphics[width=\textwidth]{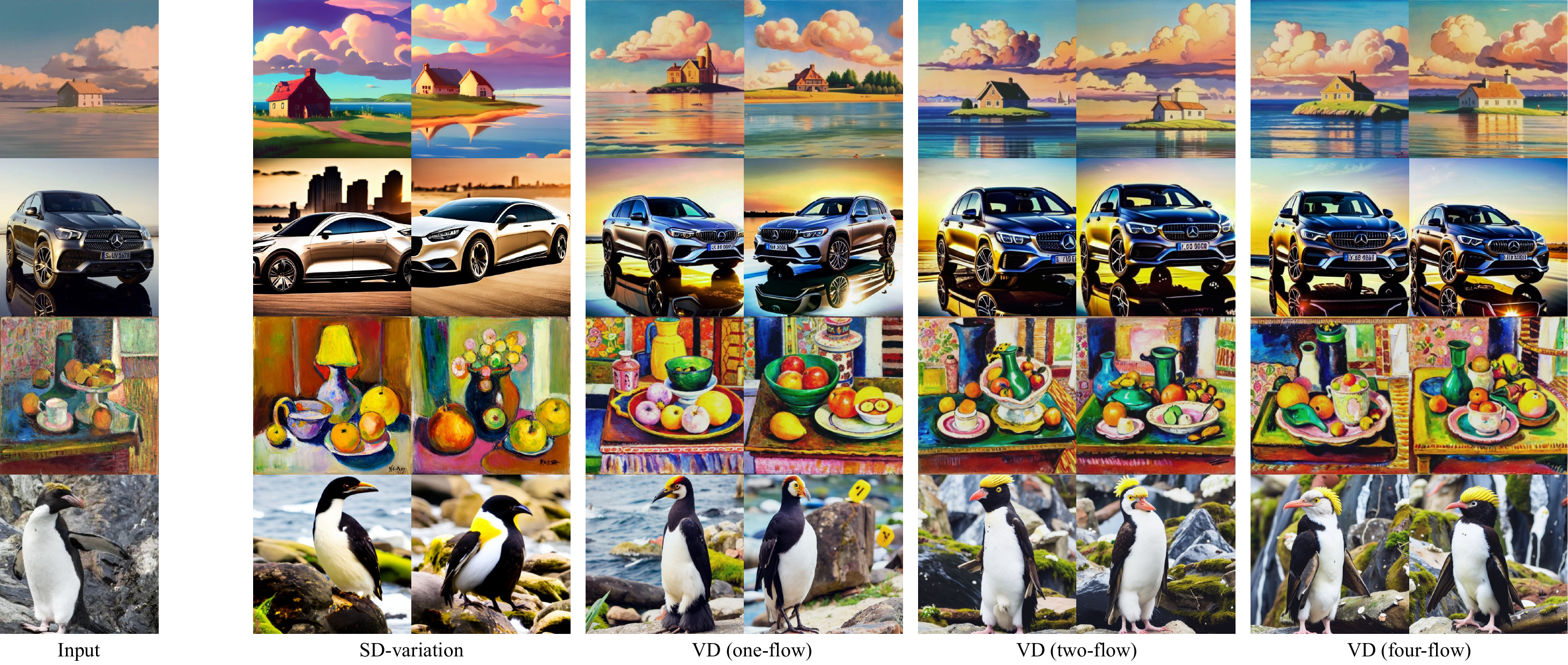}
        \caption{Image-Variation performance.}
        \vspace{0.2cm}
        \label{fig:qcompare_i2i}
    \end{subfigure}
    \begin{subfigure}[b]{0.90\textwidth}
        \centering
        \includegraphics[width=\textwidth]{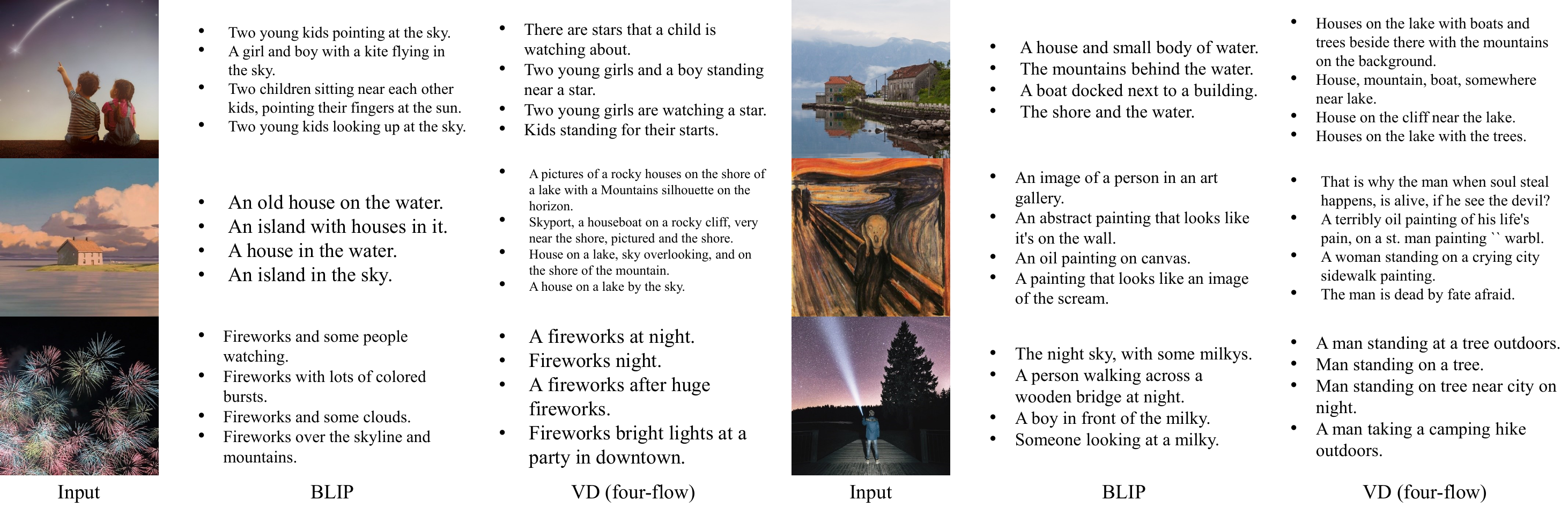}
        \caption{Image-to-Text performance.}
        \vspace{0.1cm}
        \label{fig:qcompare_i2t}
    \end{subfigure}
    
    \caption{
        These figures show the qualitative comparison between our VD models and prior works, from which we conclude that VD performs well on all three tasks. In text-to-image and image-variation, VD captures semantics from the input context more accurately. In image-to-text, VD generates more creative sentences and has a better chance to describe images with more details.
    }
    \label{fig:qcompare}
\end{figure*}

\subsection{Dataset}

We used Laion2B-en~\cite{laion} and COYO-700M~\cite{coyo} as VD's train data. Both Laion2B and COYO are collections of image-text pairs in English, in which images are collected from websites, and the corresponding captions are excerpted from HTML pages. We further filtered all data with the following criteria: a) image-text CLIP similarity scores above 0.3; b) safety scores (\ie NSWF) below 0.3; c) the probability containing watermark below 0.3; d) image aspect ratios within 0.6 to 1.6667; e) image area above $256^2 \times 0.75$. These filtered samples served as the train data for all our VD experiments. Besides, we noticed that the web crawling captions tend to be noisy, so we cleaned them with a customized algorithm described in Appendix~\ref{sec:supp_prompt_clearning}. 

\subsection{Training}

We trained VD progressively with three settings: single-flow, dual-flow, and four-flow, among which the single-flow is an image-variation model; the dual-flow is a text-to-image and image-variation model; and the four-flow is the main VD model with four tasks we majorly described in this work. During training, we kept diffusion settings close to DDPM~\cite{ddpm} and SD~\cite{ldm}, \ie, 1000 diffusion steps and linearly increasing $\beta$ from $8.5e-5$ to $1.2e-2$ according to steps. The learning rates were set to $1.e-4$ for single-flow and dual-flow, and were set to $5.e-5$ for four-flow. The single-flow model used SD checkpoint v1.4~\cite{ldm} as its initial weights, and others continued finetuning the latest checkpoint from the previous models. During training, we set different gradient scales for different layers to best cooperate with the initial weights. One can find these details in Table~\ref{table:gmsetting}. The effective batch size was 2048 for single-flow, 1024 for dual-flow, and 512 for four-flow. The logic behind the learning rates, batch sizes, and gradient scales is to roughly balance each gradient step while training. All models were trained with 30 million samples on resolution 256, followed by 6.4 million samples on resolution 512. Compared with SDv1.4, which was trained on 500 plus 230 million samples on resolutions 256 and 512, VD's training cost is more affordable, benefiting researchers in the long run.

\begin{table}[h!]
\centering
\resizebox{0.9\columnwidth}{!}{
    \begin{tabular}{
            C{1.8cm}
            C{1.2cm}
            C{1.2cm}
            C{1cm}
            C{1cm}
            C{1cm}}
        \toprule
            & Data(I) & Data(T)
            & Ctx(I) & Ctx(T)
            & Global
            \\
        \midrule
        VD (1-flow)
            & 0.1
            & --
            & 1.0
            & --
            & 0.1
            \\
        VD (2-flow)
            & 0.1
            & --
            & 1.0
            & 1.0
            & 0.1
            \\
        VD (4-flow)
            & 0.2
            & 1.0
            & 1.0
            & 1.0
            & 0.1
            \\
        \bottomrule
    \end{tabular}
}
\vspace{0.3cm}
\caption{
    This table shows the gradient scales used by different layers when training various settings of VD. Data(I) means the image data layer, so on and so forth.
}
\label{table:gmsetting}
\end{table}

\subsection{Performance}

To the best of our knowledge, VD is the first image-text multi-flow multimodal model that can be evaluated across different tasks. Thus, we chose single-task-focused prior works as our baselines when comparing the performance. Explicitly speaking: we chose SDv1.4~\cite{ldm} as our text-to-image baseline; SD-variation~\cite{sd-justin} (\ie a finetuned SD for image-variation) as our image-variation baseline; and BLIP~\cite{blip} as our image-to-text baseline. We conducted both qualitative and quantitative comparisons between baselines and various versions of VD, \ie, dual-flow and four-flow for text-to-image, and all three models for image-variation. Although DALLE2~\cite{dalle2} and Imagen~\cite{imagen} also achieved SOTA on text-to-image, they were not compared because of no publicly available code and model. For image-to-text (\ie image captioning), we only compare BLIP~\cite{blip} with our four-flow VD since other settings do not support this task.

\begin{figure}[t]
\centering
\begin{subfigure}[b]{0.475\columnwidth}
    \resizebox{\columnwidth}{!}{
        \begin{tikzpicture}
            \begin{axis}[
                major x tick style = transparent,
                ybar,
                bar width=8pt,
                ymajorgrids = true,
                width=8cm,height=6cm,
                title=Text-to-Image,
                ylabel = Count,
                symbolic x coords={A,B,C,D},
                xtick = data,
                scaled y ticks = false,
                ymin = -40,
            ]
                \addplot[style={blue,fill=blue,mark=none}]
                    coordinates {(A,72) (B,131) (C,79) (D,128)};
        
                \addplot[style={cyan,fill=cyan,mark=none}]
                    coordinates {(A,92) (B,156) (C,35) (D,183)};
        
                \addplot[style={gray,fill=gray,mark=none}]
                    coordinates {(A,336) (B,213) (C,386) (D,189)};
                \end{axis}
        \end{tikzpicture}
    }
\end{subfigure}
\hfill
\begin{subfigure}[b]{0.475\columnwidth}
    \resizebox{\columnwidth}{!}{
        \begin{tikzpicture}
            \begin{axis}[
                major x tick style = transparent,
                ybar,
                bar width=8pt,
                ymajorgrids = true,
                width=8cm,height=6cm,
                title=Image-Variation,
                ylabel = Count,
                symbolic x coords={A,B,C,D},
                xtick = data,
                scaled y ticks = false,
                ymin = -40,
            ]
                \addplot[style={blue,fill=blue,mark=none}]
                    coordinates {(A,23) (B,25) (C,2) (D,10)};
        
                \addplot[style={cyan,fill=cyan,mark=none}]
                    coordinates {(A,406) (B,336) (C,485) (D,408)};
        
                \addplot[style={gray,fill=gray,mark=none}]
                    coordinates {(A,71) (B,139) (C,13) (D,82)};
                \end{axis}
        \end{tikzpicture}
    }
\end{subfigure}
\caption{User studies on text-to-image and image-variation in which we count the votes from 4 individual moderators on SD (\textcolor{blue}{blue}), VD (\textcolor{cyan}{cyan}), or equally good (\textcolor{gray}{gray}).}
\vspace{-0.2cm}
\label{fig:user_study}
\end{figure}
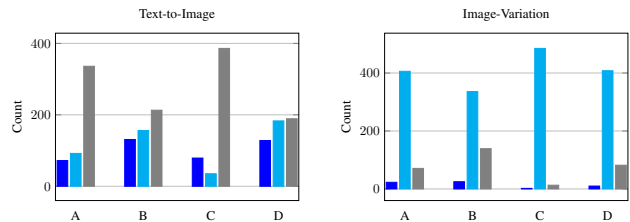

Figure~\ref{fig:qcompare} compares VD's qualitative performance with its baseline, in which images in each row are created with the same random seeds for better quality checks. We also compute text-to-image and image-variation FID scores by comparing 30000 randomly generated samples with the validation set of COCO-caption~\cite{coco}. In Figure~\ref{fig:quant_result}, we list VD's performance along with other related works. We also plot the changes in VD's FID according to the unconditional guidance scale (\ie the classifier-free guidance scale). Lastly, we carried out user studies on 2000 samples from COCO-Caption~\cite{coco} split by four moderators, in which moderators were asked to vote for better quality or ``equally good" (see Figure~\ref{fig:user_study}).

\begin{figure}[t]
\centering
\begin{subfigure}[b]{0.475\columnwidth}
    \resizebox{\columnwidth}{!}{
    \begin{tabular}{
            C{4cm}
            C{2cm}}
        \toprule
            Method & FID $\downarrow$
            \\
        \midrule
        \midrule
            \multicolumn{2}{c}{(A) Text-to-Image Synthesis}
            \\
        \midrule
            CogView~\cite{cogview} & 27.10 \\
            LAFITE~\cite{lafite}  & 26.94 \\
            GLIDE~\cite{glide}   & 12.24 \\
            Make-a-Scene~\cite{mask-a-scene} & 11.84 \\
            LDM~\cite{ldm} & 12.63 \\
            SD (baseline) & 11.21 \scriptsize{$\pm$0.03} \\
            \textbf{VD (four-flow)} & \textbf{11.10 \scriptsize{$\pm$0.09}} \\
        \midrule
        \midrule
            \multicolumn{2}{c}{(B) Image-Variation Synthesis}
            \\
        \midrule
            SD (baseline) & 18.81 \scriptsize{$\pm$0.06} \\
            \textbf{VD (four-flow)} & \textbf{4.57 \scriptsize{$\pm$0.02}} \\
        \bottomrule
    \end{tabular}
}

\end{subfigure}
\hfill
\begin{subfigure}[b]{0.475\columnwidth}
    \resizebox{\columnwidth}{!}{
    \begin{tikzpicture}
    \begin{axis}[
        xlabel=Unconditional Guidance Scale,
        ylabel=FID,
        xmin=2,
        xmax=8,
        ymin=4,
        ymax=14,
        xtick={2.5,5.0,7.5},
        width=8cm,height=8.2cm,
        legend style={at={(0.22,0.96)},anchor=north,legend cell align=left},
    ]
    \addplot+[
        blue,
        mark=*, mark size=3pt, mark options={fill=blue},
        error bars/.cd,y dir=both,y explicit,
    ] coordinates {
        (2.5, 11.3355988257) +- (0, 0.0796762530121)
        (5.0, 11.1023553055) +- (0, 0.0916626027159)
        (7.5, 13.405008415)  +- (0, 0.0681356441779)
    };\label{t2i-fid}\addlegendentry{FID(T2I)}
    \addplot+[
        cyan,
        mark=*, mark size=3pt, mark options={fill=cyan},
        error bars/.cd,y dir=both,y explicit,
    ] coordinates {
        (2.5, 4.56646982092) +- (0, 0.0211169100996)
        (5.0, 6.7336671616) +- (0, 0.0299270554132)
        (7.5, 8.48817909203)  +- (0, 0.0205315863737)
    };\label{iv-fid}\addlegendentry{FID(IV)}
    \end{axis}
    \end{tikzpicture}
}
    \vspace{-0.9cm}
\end{subfigure}
\caption{FID scores of VD comparing with baseline and prior approaches, and under various unconditional (classifier-free) guidance scales.}
\label{fig:quant_result}
\end{figure}
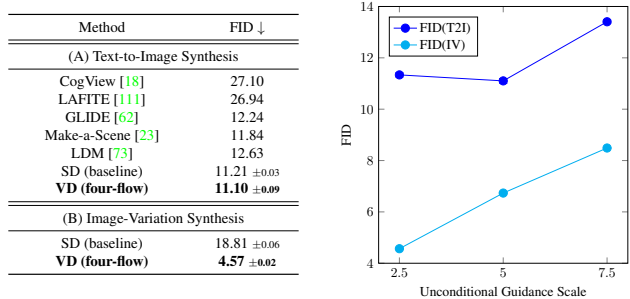

Through all results, we not only demonstrated that VD outperforms its baseline on these primary tasks, but reveals the effectiveness of our multi-flow multimodal diffusion framework in which context and data with distinct modalities can be analyzed and generated in one unified model.

\subsection{Disentanglement of style and semantic}

One exciting discovery of our VD is that it can enhance or reduce image styles from semantics without further supervision. Such a phenomenon inspires us to explore a novel area where disentanglement between styles and semantics can happen on images with arbitrary contents in arbitrary styles. Recall that prior works such as~\cite{gan_dissect, ganspace} explored similar properties in GAN latent spaces, but their domain of study was restricted to well-aligned data such as faces or churches. To our best knowledge, we are the first group exploring: a) unsupervised semantic and style disentanglement on natural images without domain specifications; b) semantic and style disentanglement on diffusion models' latent space.

\begin{figure}[t]
    \centering
    \includegraphics[width=0.48\textwidth]{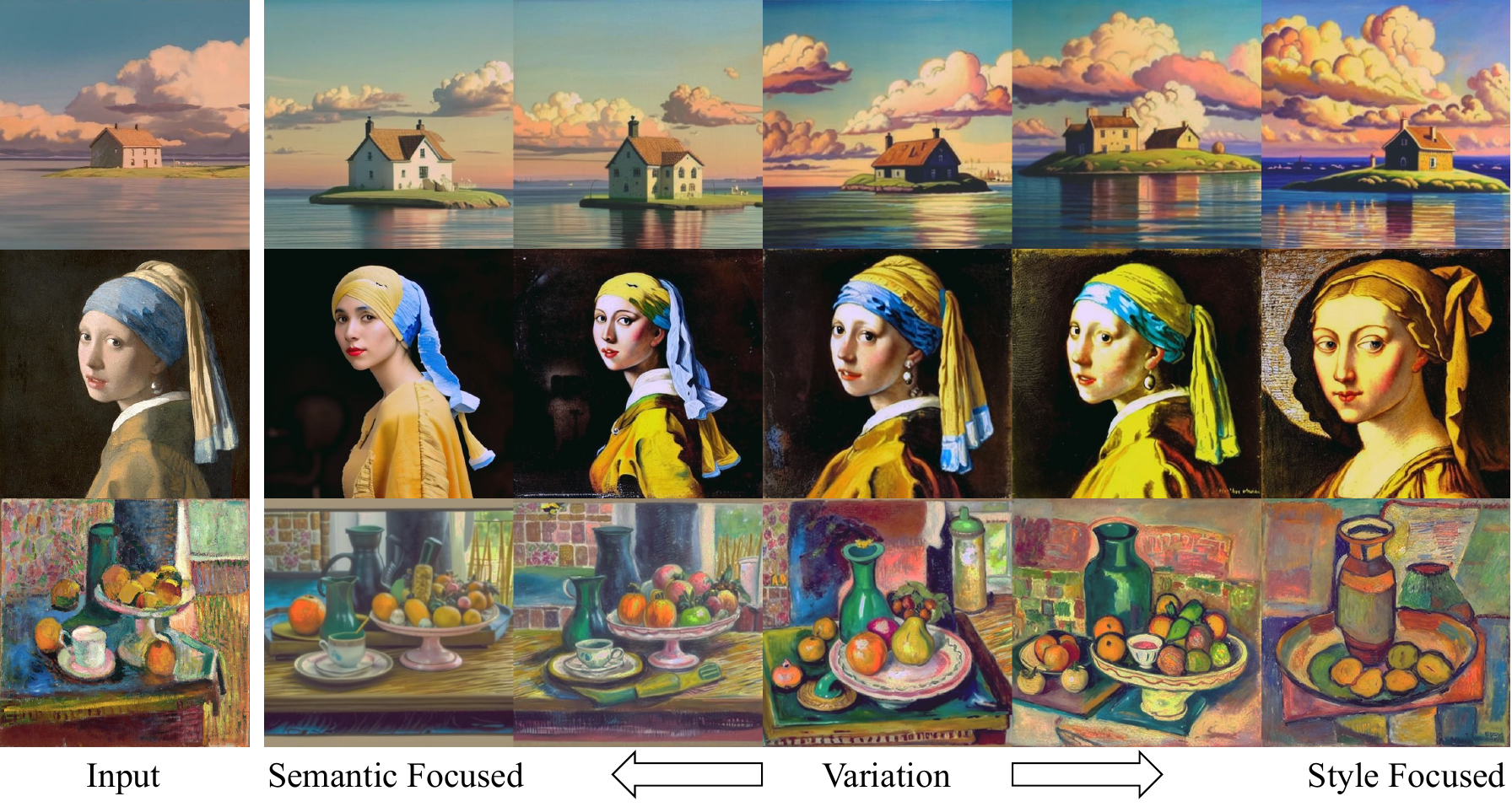}
    \vspace{-0.4cm}
    \caption{Our VD can disentangle image semantics from styles and vice versa. In this figure, we first generate variations of the input images and then manipulate them focused on either semantics (to the left) or styles (to the right).}
    \label{fig:disentanglement}
\end{figure}

Figure~\ref{fig:disentanglement} shows the disentanglement results of VD. In practice, we notice that both two-flow and four-flow models serve similar performance, while single-flow has slightly lower performance. This may be due to the caption-agnostic and insufficient training that reduced the model's capacity. More details and analysis can be found in Appendix~\ref{sec:supp_disentanglement}.

\begin{figure}[t]
    \centering
    \includegraphics[width=0.48\textwidth]{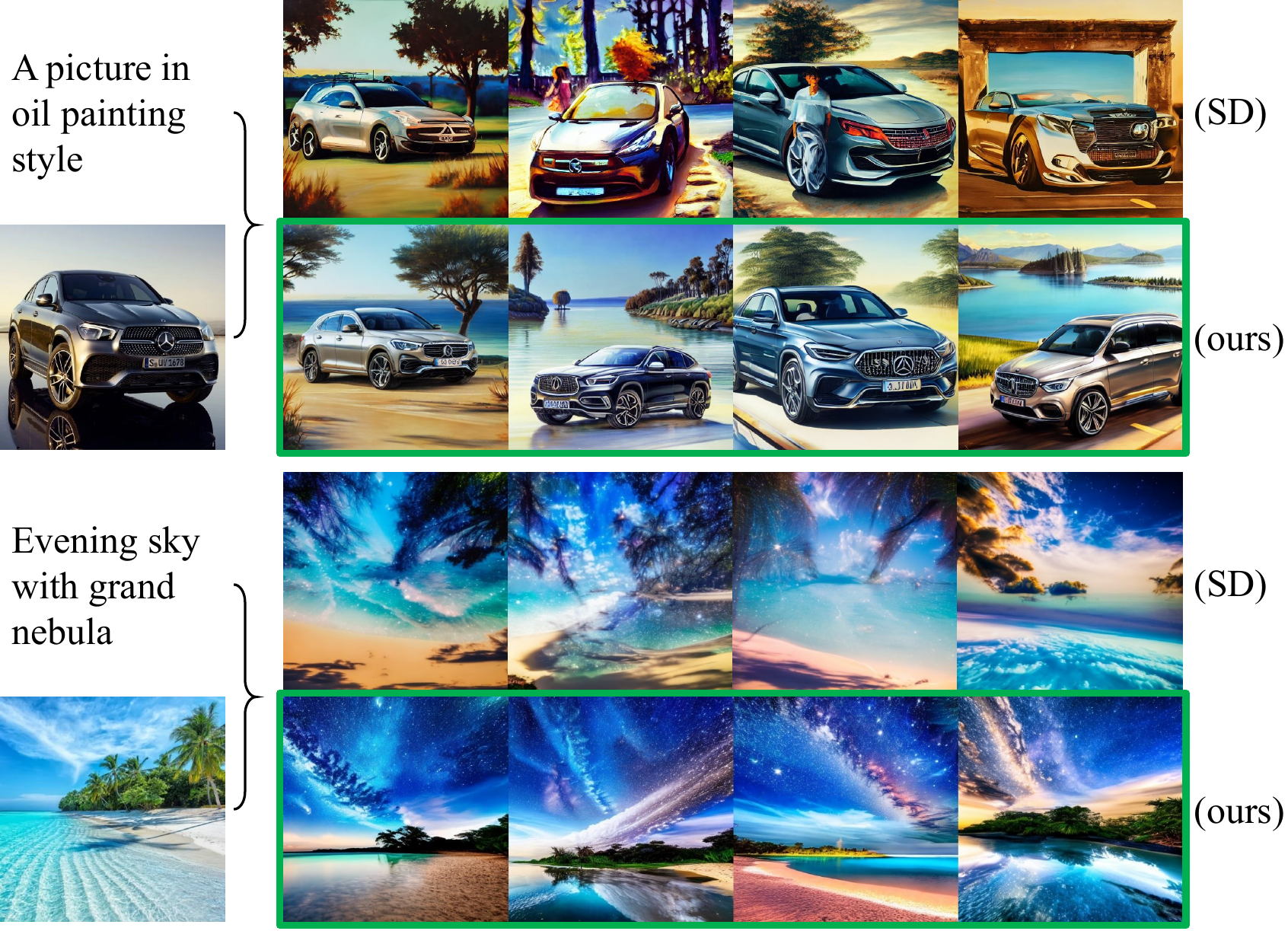}
    \vspace{-0.2cm}
    \caption{This figure shows images generated from dual-context blender (one image and one prompt). Images without borders are baseline results generated by ensembling SDv1.4~\cite{ldm} with SD-variation~\cite{sd-justin}. Images with green borders are VD's outputs (ours) with a deeper level of mixing. To fairly compare the performance, samples in the same columns use the same random seed and initial noise inputs.}
    \label{fig:dual_guidance}
\end{figure}

\subsection{Dual- and multi-context blender}

Since VD is a unified model for multiple tasks, generation from multi-context becomes a natural extension for VD. Recall that a baseline multi-context generation can be achieved by mixing up diffusion steps from distinct models~\cite{dm_morecontrol}. However, in practice, we notice such a baseline cannot reach satisfactory results despite doubling the model usage. Figure~\ref{fig:dual_guidance} compares the dual-context results using one text and one image, in which we use the mixing of SDv1.4~\cite{ldm} (text-to-image) and SD-variation~\cite{sd-justin} (image-variation) as our baseline (labeled as SD). One may easily notice that VD generates more natural-looking results with fewer distortions. We believe that the good performance of VD is largely attributed to its multi-flow structure, through which intermediate features generated from different contexts can be merged on a much deeper level (\ie layer-level or attention-level), instead of merged on the shallow model-level between diffusion steps. More details regarding mixing levels can be found in Appendix~\ref{sec:supp_dcg}.

\begin{figure}[t]
    \centering
    \includegraphics[width=0.46\textwidth]{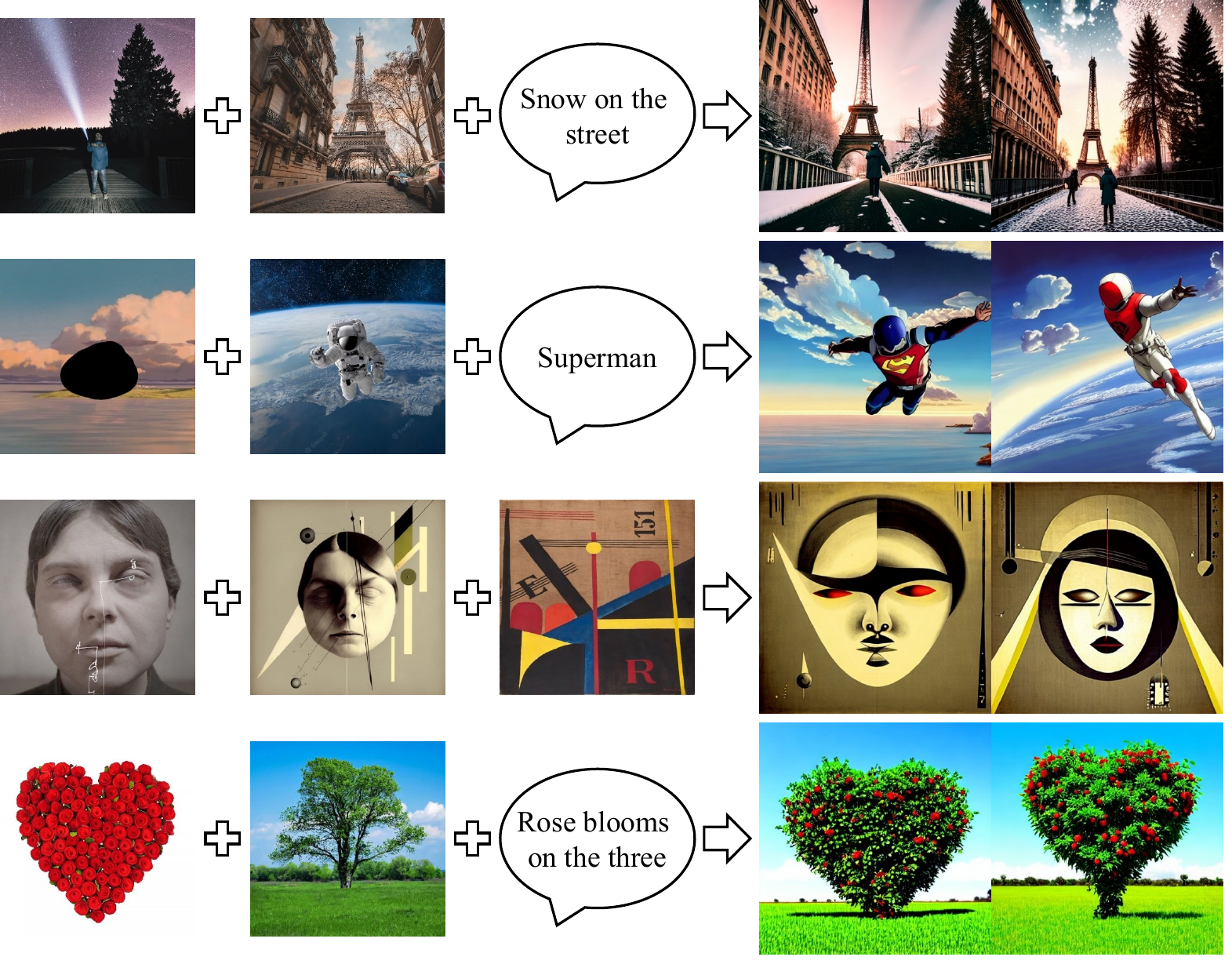}
    \caption{This figure shows images created with VD's multi-context blender in which multiple images with optional text and masks are applied as contexts. One can notice that VD can smoothly transfer and reconstruct semantic from contexts to outputs.}
    \label{fig:multi_context}
\end{figure}

We further expand this task to a more generalized form with multi-context, resulting in the multi-context blender application. The multi-context blender for VD supports an optional text context, several image contexts, and optional image masks in order to guide the generation process with more detail controls. Figure~\ref{fig:multi_context} shows the performance of our multi-context blender. Notice that there are other recent works such as~\cite{p2p, insp2p, controlnet, attend_and_excite, dreambooth, custom_diffusion, imagic} focused on the broader image editing topic. We encourage readers to check our Appendix~\ref{sec:supp_dcg} and~\ref{sec:supp_mcg} for more details and comparisons. 

\vspace{-0.1cm}
\section{Conclusion}
In this article, we proposed Versatile Diffusion that handles text, image, and variations all in one, from which we generalized a multi-flow multimodal framework that can further extend to new tasks and domains. Through inclusive experiments, we demonstrate that such a multi-flow multimodal diffusion method can perform well on both primary tasks and applications. Moreover, VD can be a heuristic step toward universal AI research.


{\small
\bibliographystyle{ieee_fullname}
\bibliography{egbib}
}

\clearpage

\onecolumn
\begin{appendices}
\vspace{1cm}

\section{Application Details}

\subsection{Disentanglement of Style and Semantic}
\label{sec:supp_disentanglement}

\begin{figure}[t]
    \centering
    \includegraphics[width=0.9\textwidth]{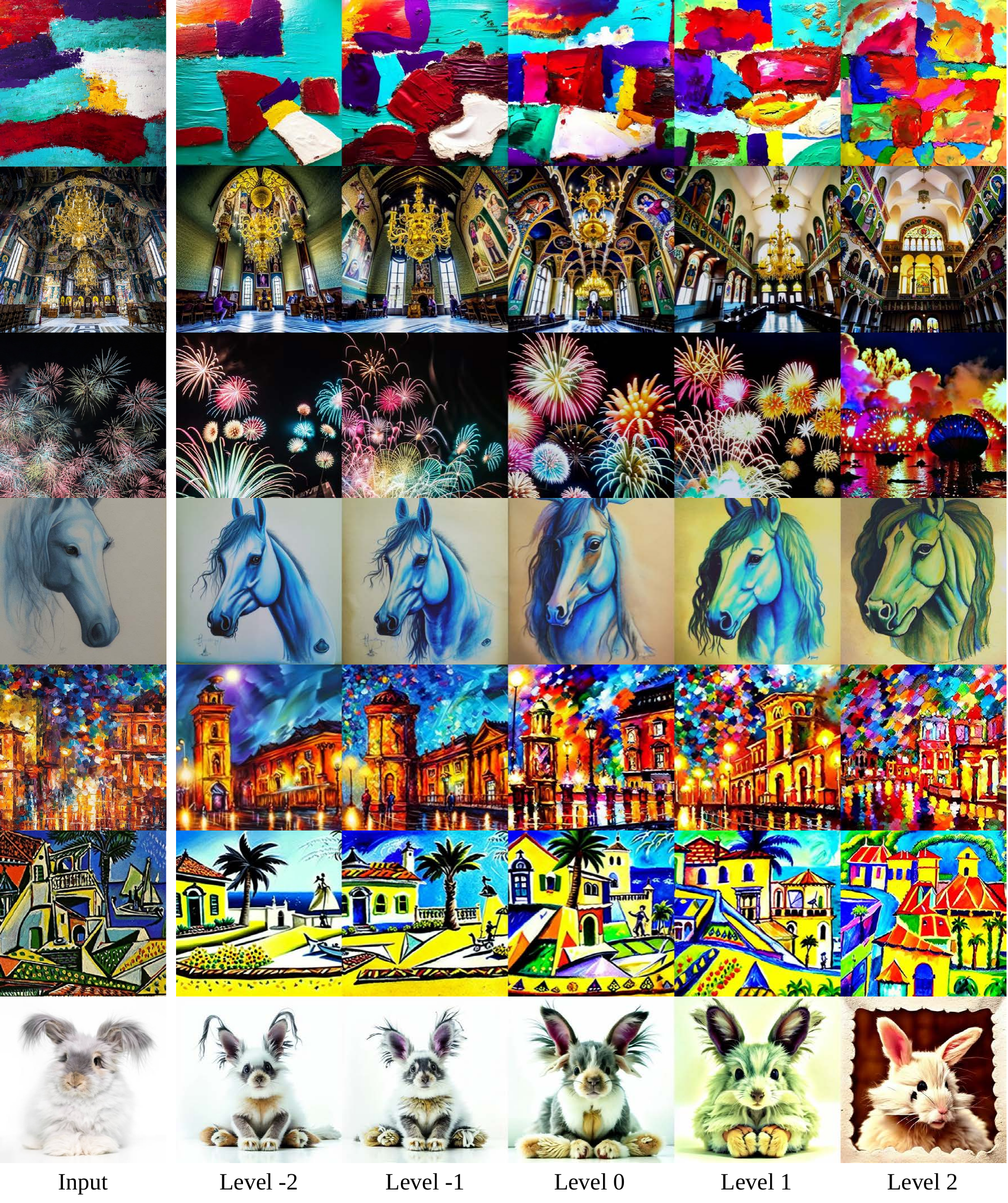}
    \caption{Additional figures that show the performance of our proposed disentanglement application with different levels.}
    \vspace{0.2cm}
    \label{fig:supp_disentanglement}
\end{figure}

The disentanglement application conducts controllable image-variation, supported by the image-variation flow of VD. Such flow consists of AutoKL, CLIP image encoder, and VD's diffuser with image data layers and image context layers. The core strategy of the disentanglement is to manipulate the 257$\times$768 CLIP image context embedding, which guided the diffusion process via cross-attention. Recall that these embeddings are generated by the visual transformer~\cite{vit}, which begins with one global feature vector followed by 256 local feature vectors corresponding to image local patches. We first split the vector into the single global vector and the following 256 local vectors. We keep the global vector untouched and compute the principal components from the rest of the feature vectors. When manipulating the context embeddings, we notice that the first couple of principal components (\ie the major principal components of the matrix) hold the style information (\ie color, art, stroke styles), and the remaining principal components hold the semantic information (\ie, objects, object locations, identity). Thus, in practice, we generated image variations with style focuses from the guidance of the low-rank context embedding that hosts only major principal components. And we generated image variations with semantic focuses when we removed these major principal components from context embeddings. In Figure~\ref{fig:supp_disentanglement}, we show additional qualitative results, in which we standardize the disentanglement with five levels: 

\setlist[enumerate,1]{leftmargin=1.5cm, rightmargin=1.5cm}
\begin{enumerate}[label=(\alph*)]
    \item 0 represents normal image-variation.
    \vspace{-0.5em}
    \item -1 and -2 are semantic-focused by removing one and two major principal components.
    \vspace{-0.5em}
    \item 1 and 2 are style focuses results corresponding to keeping only 10 and 2 major principal component.
\end{enumerate}

\begin{figure}[t]
    \centering
    \includegraphics[width=0.8\textwidth]{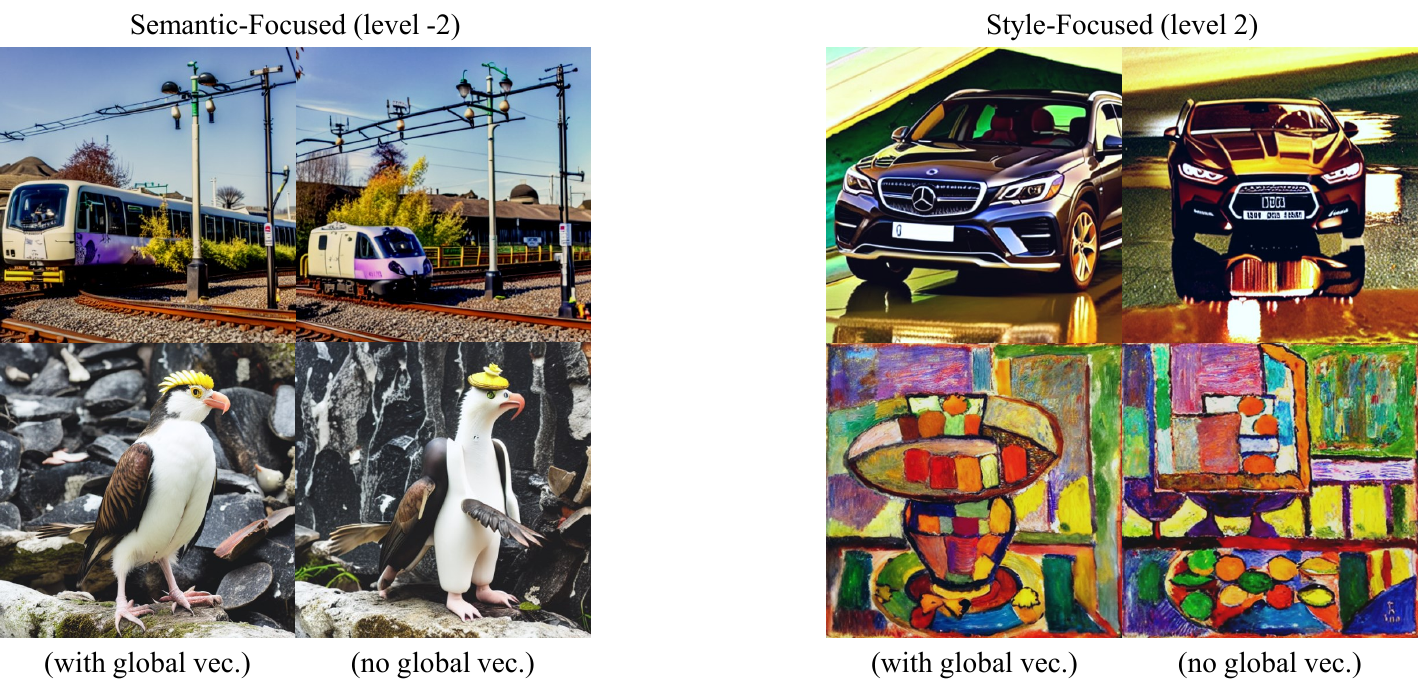}
    \caption{This figure shows four comparisons between samples generated via context with the global vector and without the global vector. The two left cases are semantic-focused outputs, and the two right cases are style-focused outputs. }
    \vspace{0.2cm}
    \label{fig:supp_disentanglement_gvec}
\end{figure}

In practice, we also notice that principal components after order 50 have little effect on results. Therefore, we can speed up the disentanglement PCA by just computing the first 50 principal components and then conducting the manipulation. For the global feature vector, we notice that it mainly serves as a semantic feature that controls object location information. Hence, removing it may negatively impact image structure (see Figure~\ref{fig:supp_disentanglement_gvec}) but may be useful in some art generation cases. We encourage researchers to explore further the low-rank subspace of the CLIP Image embedding for more exciting applications.

\subsection{Dual-context Blender}
\label{sec:supp_dcg}

\begin{figure}[t]
\centering
    \begin{subfigure}[b]{0.48\textwidth}
        \centering
        \includegraphics[width=\textwidth]{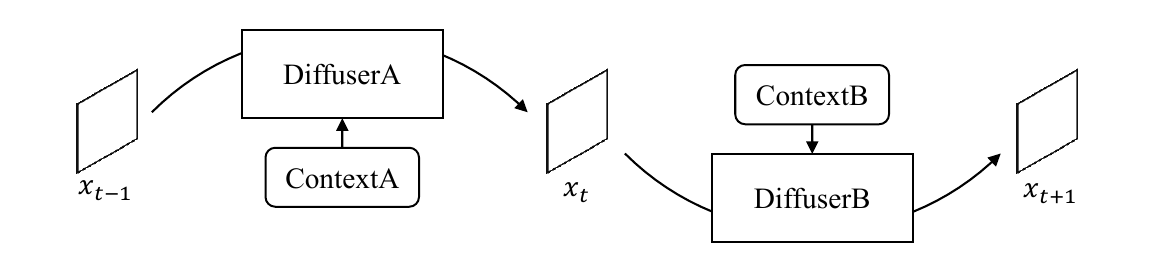}
        \subcaption{Model-level Mixing A}
        \vspace{0.5cm}
    \end{subfigure}
    \begin{subfigure}[b]{0.48\textwidth}
        \centering
        \includegraphics[width=\textwidth]{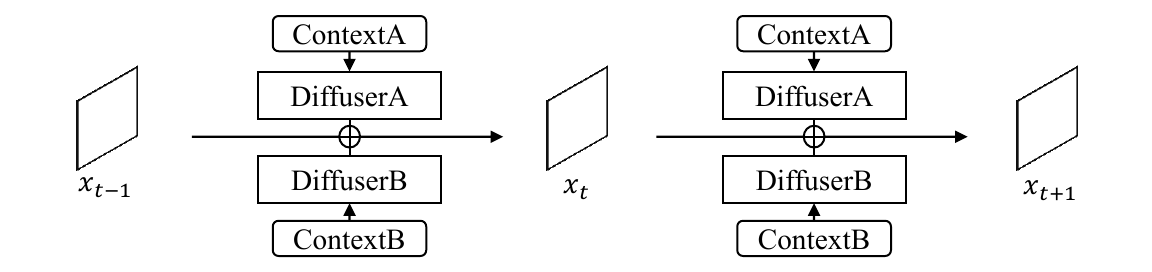}
        \subcaption{Model-level Mixing B}
        \vspace{0.5cm}
    \end{subfigure}
    \\
    \begin{subfigure}[b]{0.48\textwidth}
        \centering
        \includegraphics[width=\textwidth]{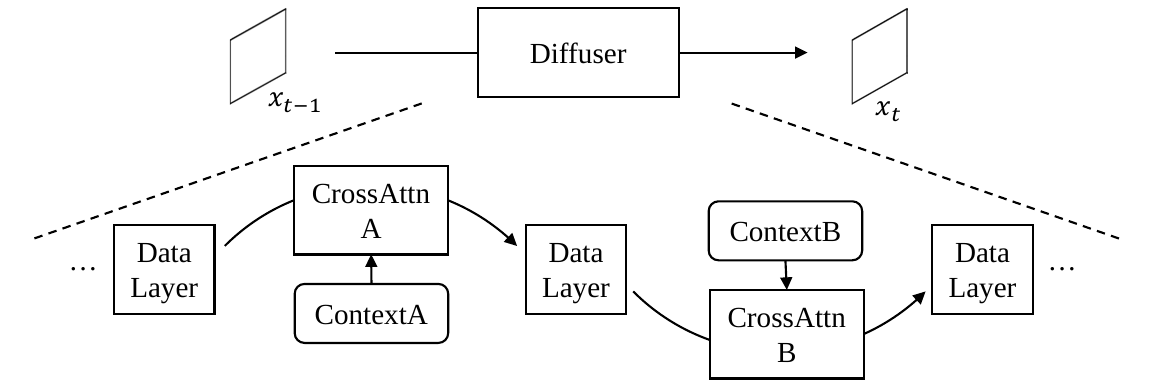}
        \subcaption{Layer-level Mixing}
        \vspace{0.5cm}
    \end{subfigure}
    \begin{subfigure}[b]{0.48\textwidth}
        \centering
        \includegraphics[width=\textwidth]{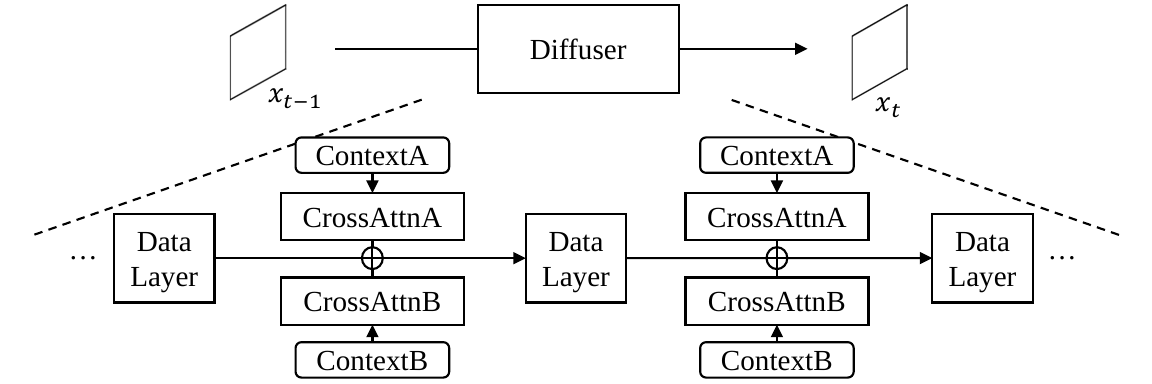}
        \subcaption{Attention-level Mixing}
        \vspace{0.5cm}
    \end{subfigure}
\caption{The graphic explanation of the three mixing strategies we mentioned in our dual-context blender. Both (a) and (b) are two types of model-level mixing.}
\label{fig:supp_mixing}
\end{figure}

The dual-context blender for VD is to generate images through the guidance of one image and one text prompt. Theoretically, such an application can also be used to create new text/sentences, but the results are less exciting than creating new images. As mentioned in the main article, the dual-context blender, or the multi-context blender, can be achieved by ensembling two models in which we mix the diffusion steps from one model after another (type A), or weighted sum up both models' outputs (type B). However, in practice, we notice that such approaches may cause structure distortions and highlight wrong semantics despite doubling the model usage. Unlike these simple ensembling methods, VD can carry out this task via a much deeper level of mixing due to its multi-flow multimodal framework. As mentioned in the main article Section 3.2, our framework has three layer groups: global, data, and context. When generating images, features diffuse through the shared data layers (\ie ResBlocks), and then mix up via different context layers with two options: layer-level mixing or attention-level mixing. In layer-level mixing, we diffuse features through different context layers (\ie cross-attention) that follow a preset schedule. For example, we diffuse features through one image cross-attention, then a text cross-attention, \etc. In attention-level mixing, both immediate features after context layers are included via a weighted sum and then passed to the network's next block (See Figure~\ref{fig:supp_mixing}). 

\begin{figure}[tp]
    \centering
    \includegraphics[width=0.8\textwidth]{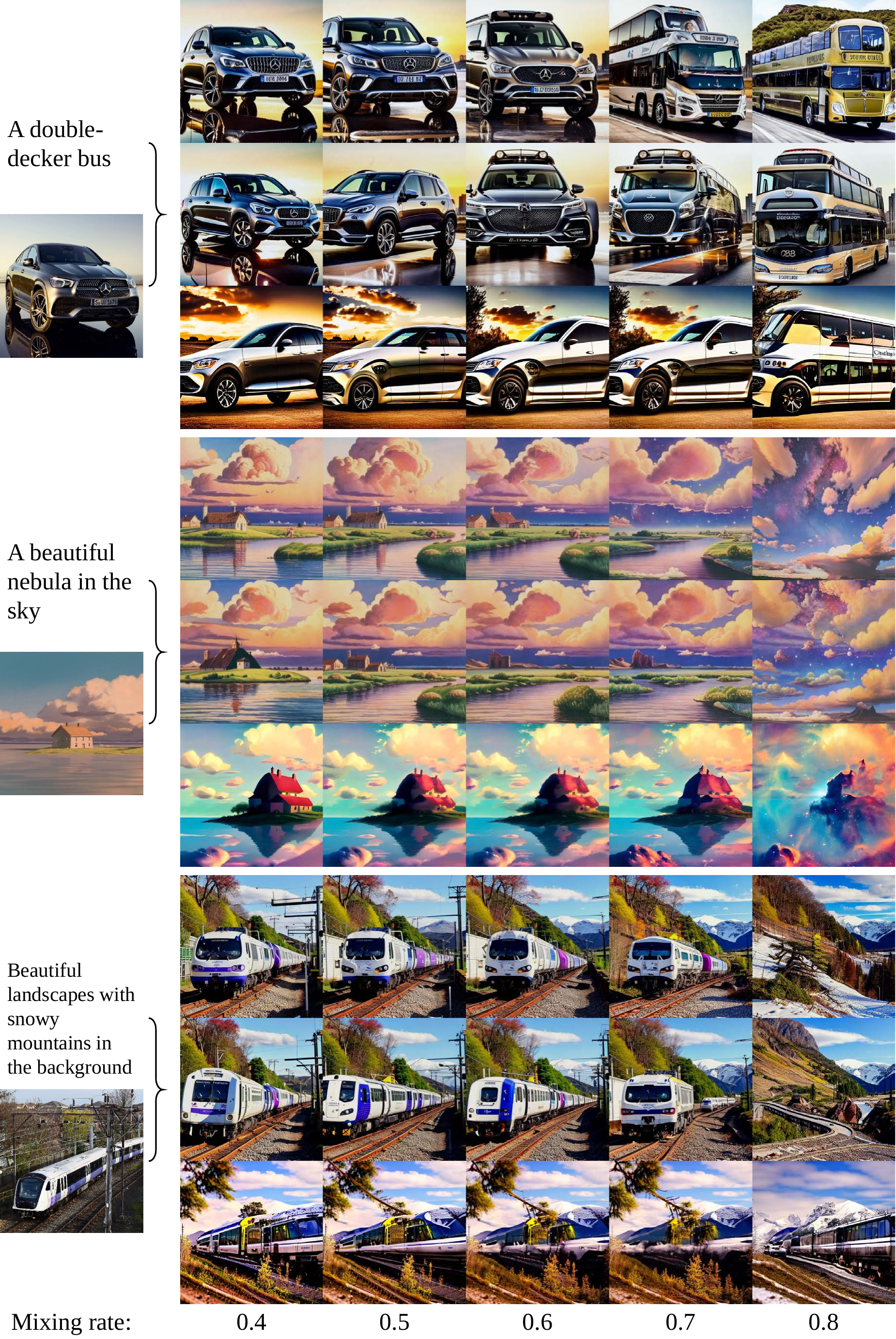}
    \caption{Additional figures that show the performance of our proposed dual-context blender. The horizontal axis shows the mixing rate we use in these samples: small mixing rates lead to image-focused (to the left), and large mixing rates lead to text-focused (to the right). For each sample, we show three rows, from top to bottom are the results of attention-level mixing, layer-level mixing, and model-level mixing, in which the top row (attention-level mixing) is the best.}
    \vspace{0.2cm}
    \label{fig:supp_dual_guided}
\end{figure}

We believe that the success of the dual-context blender heavily relies on VD's multi-flow design that can merge contexts in a harmonized way. An example shows in Figure~\ref{fig:supp_dual_guided} in which we generate an image using \textit{a car} as image context and a prompt \textit{a double-decker bus} as text context. Such a case manually brings challenges in mixing, since a Benz car in the image and a double-decker bus described in the prompt have completely different shapes. However, attention-level mixing nicely resolves such conflict and we notice a smooth transition between these two contexts with increasing mixing rates. On the other hand, our results indicate that layer-level mixing slightly underperforms attention-level mixing as the generated vehicle shows some noticeable distortion (see the wheels in the second line). Lastly, we show the results from model-level mixing using two of our baseline models SDv1.4 and SD-variation, which perform the worst among all three methods. Given these results, we conclude that VD is critical for the success of our dual-context blender tasks, in which its multi-flow multimodal network framework is the key to effectively resolving potential conflict and merging various contexts.

\subsection{Multi-context Blender with Optional Masks}
\label{sec:supp_mcg}

Multi-context blender is an extension of our dual-context blender with the following changes: 

\setlist[enumerate,1]{leftmargin=1.5cm, rightmargin=1.5cm}
\begin{enumerate}[label=(\alph*)]
    \item It takes more than one image as a concatenated context of image type.
    \vspace{-0.5em}
    \item It accepts additional scale control on each of the input images.
    \vspace{-0.5em}
    \item It allows adding individual masks to precisely control the generated output based on reference images.
\end{enumerate}

Adding an extra image as reference context is actually simpler than adding an extra context type. Specifically for VD, context from multiple images can be concatenated, forming a more extended sequence of context embedding that later serves as input to content layers. For example, context embedding one image is a $257\times 768$ embedding in which $257$ is the number of embedding vectors and $768$ is the embedding channel. Context embedding for two images is simply a concatenated feature of two images embedding along the number dimension, making it $514\times 768$, and so-on-so-forth for more images. Therefore, such operations do not alter any mixing strategies that we used in the dual-context blender.  

\begin{figure}[h]
    \centering
    \vspace{0.5cm}
    \includegraphics[width=0.8\textwidth]{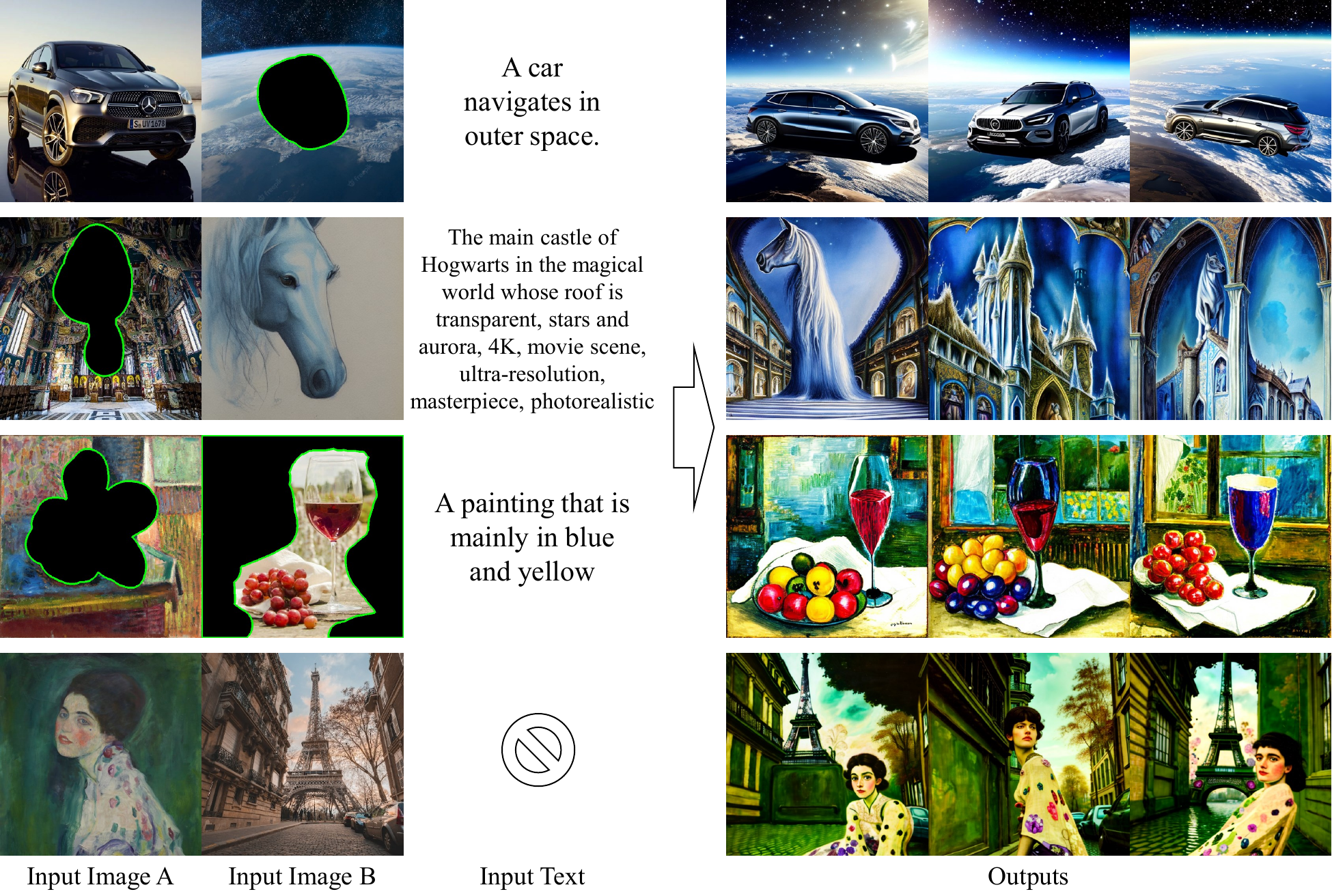}
    \vspace{0.2cm}
    \caption{Additional examples generated by VD's multi-context blenders. As mentioned in the paper, our multi-context blender uses multiple images with optional masks adding an extra text prompt (also optional) to guide its generation process. Its outputs are then a blender of all input context.}
    \label{fig:supp_more_tcg}
\end{figure}

To make more precise control of images used in our multi-context blender, we involved two controllable parameters: image scales and image masks. Image scales are simple multipliers associated with underlining image context embeddings, while image masks involve more complex designs. We notice that a naive solution to replace contents in masks with zeros may confuse the model of generating images with black patches. Thus we altered the CLIP network, in which the raw image features after the first convolution projection of ViT~\cite{vit} and the input positional encodings are filled with zeros according to masks before inputting the transformers. As a result, we successfully involved scales and masks in our blender application. More results can be found in Figure~\ref{fig:supp_more_tcg}.

\subsection{Editable I2T2I}
\label{sec:supp_i2t2i}

Since VD supports both image-to-text and text-to-image, one heuristic image editing approach we can do is to edit images with the following steps: \textbf{(a)} \textit{convert image to text}, \textbf{(b)} \textit{edit text}, and \textbf{(c)} \textit{convert text back to the image}. We named this approach \textit{image-to-text-to-image} \textbf{(I2T2I)}. Although the principle of I2T2I is simple, we notice that in practice, many issues may negatively affect the outputs, making them less robust than expected. 

\begin{figure}[h]
    \centering
    \vspace{0.5cm}
    \includegraphics[width=0.8\textwidth]{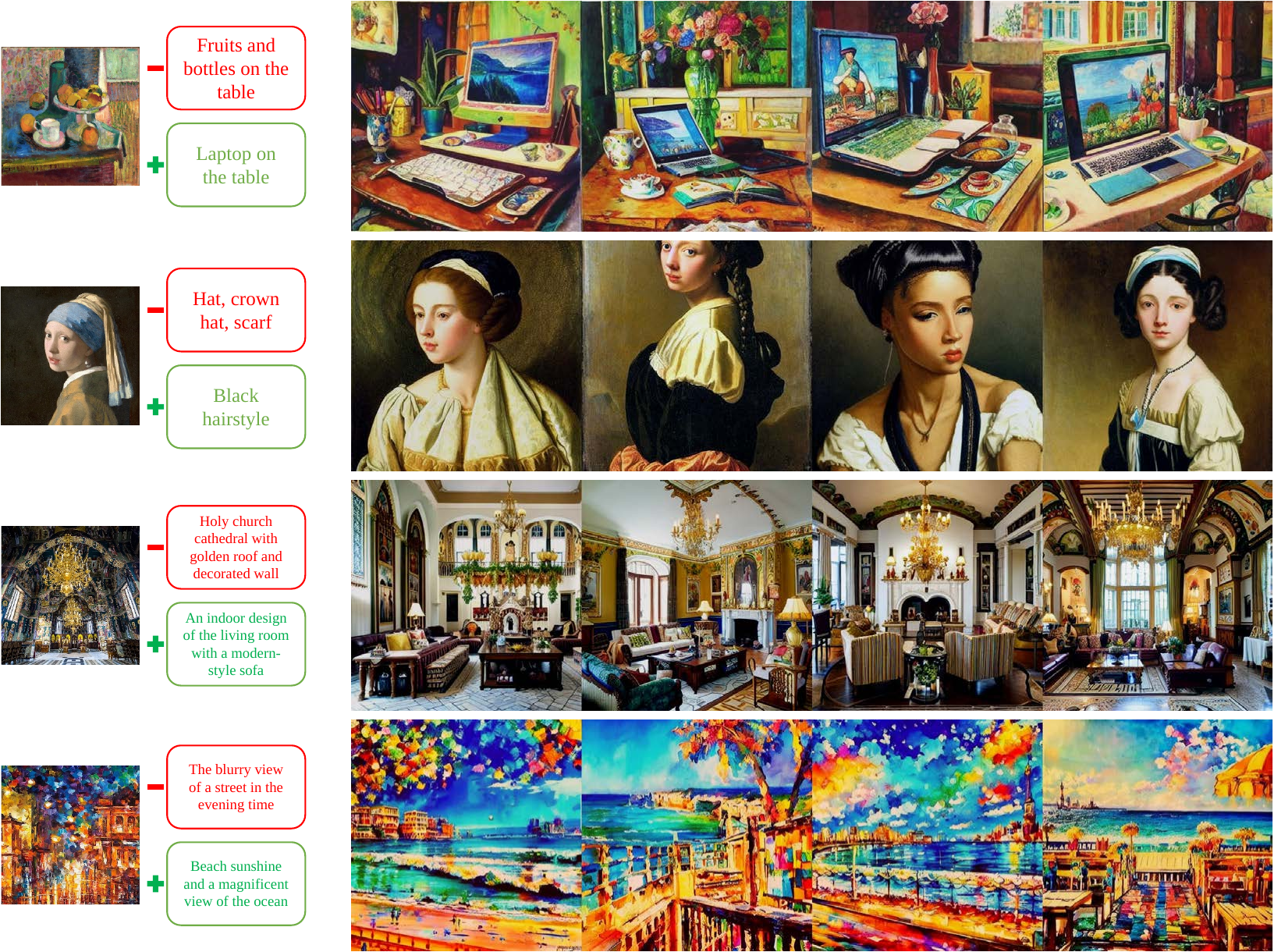}
    \vspace{0.2cm}
    \caption{This figure shows the performance of our proposed image editing method I2T2I with negative and positive prompts. (see Section~\ref{sec:supp_i2t2i})}
    \label{fig:supp_i2t2i}
\end{figure}

Unlike inpainting or multi-context blending, I2T2I requires no object masks because
one design goal of I2T2I is to let it automatically locate and substitute objects following the prompt instruction. Meanwhile, I2T2I’s output images do not match its input images pixel by pixel, which is a result of semantic distillation and content creation. Figure~\ref{fig:supp_i2t2i} shows the prototype results of our I2T2I in which old contents inside the image are removed and re-placed via prompt editing. To the best of our knowledge, this is the first attempt at creating and editing images by combining image-to-text, text editing, and then text-to-image. Yet we notice the following issues that may dramatically decrease the performance: 

\setlist[enumerate,1]{leftmargin=1.5cm, rightmargin=1.5cm}
\begin{enumerate}[label=(\alph*)]
    \item The output quality is affected by both image-to-text and text-to-image performance. The failure of either one of the sub-procedures will result in unsatisfactory results.
    \vspace{-0.5em}
    \item Sometimes, direct editing of the generated text could be infeasible, as there is no guarantee that the text contains the descriptions we would like to modify.
    \vspace{-0.5em}
    \item Image-to-text is a process of information distillation, while text-to-image is a process of information creation. Although such properties bring great flexibility in I2T2I editing, they may differ from general users' demands because they may like to keep more content from the reference images.
\end{enumerate}

To overcome these issues, we tackle these issues with the following solution. Instead of editing the text directly, we actually modify the latent text vectors as a solution to b) issue. Speaking with details, in our editing experiment, we prepare a negative and positive prompt to do the editing. The negative prompt describes image content that needs to be removed, and the positive prompt describes the content to add. When the text latent vector is ready (using image-to-text), before converting it into text, we project it on the normal space of the negative prompt latent vector and sum it up with the positive prompt latent vector. To further strengthen the positive prompt, we also compute its CLIP embedding and concatenate it with the modified prompt embedding, and then guide the image generation. Meanwhile, we adopt the ideas from our disentanglement and dual-context blender, in which we compute the style disentangled image context and use it as secondary guidance in the generation process with a 0.66 mixing rate. The final performance of I2T2I can be found in Figure~\ref{fig:supp_i2t2i}.

\newpage

\section{Image-Variation Analysis and Beyond}

\subsection{Unconditional Guidance}
\label{sec:supp_iv_ug}

We also did an in-depth investigation on the influence of the unconditional guidance (\ie classifier-free guidance) of the image-variation. Recall that such mechanism is first involved in class-guided generation~\cite{ddpm, ldm} and later largely adopted by text-to-image models such as~\cite{ldm, dalle2, imagen}. Here we recap the core math in Equation~\ref{eq:cfree_guide}:

\begin{equation}
    y = y_u + (y_c - y_u) * s,\quad y_u = G(c_{\text{uncond}}),\quad y_c = G(c_{\text{cond}}) 
\label{eq:cfree_guide}
\end{equation}

\noindent in which $y$ is the final output, $s$ is the unconditional guidance scale, $y_u$ is the unconditional output from generator $G$ using unconditional context $c_{\text{uncond}}$, and $y_c$ it the conditional output from $G$ and $c_{\text{cond}}$. For text-to-image, $c_{\text{uncond}}$ are usually set to the text embeddings encoded from empty strings. For image-variation, we have two options: 

\setlist[enumerate,1]{leftmargin=1.5cm, rightmargin=1.5cm}
\begin{enumerate}[label=(\alph*)]
    \item CLIP embeddings of empty images with all zeros
    \vspace{-0.5em}
    \item All-zero embeddings
\end{enumerate}

\begin{figure}[h]
    \centering
    \includegraphics[width=0.96\textwidth]{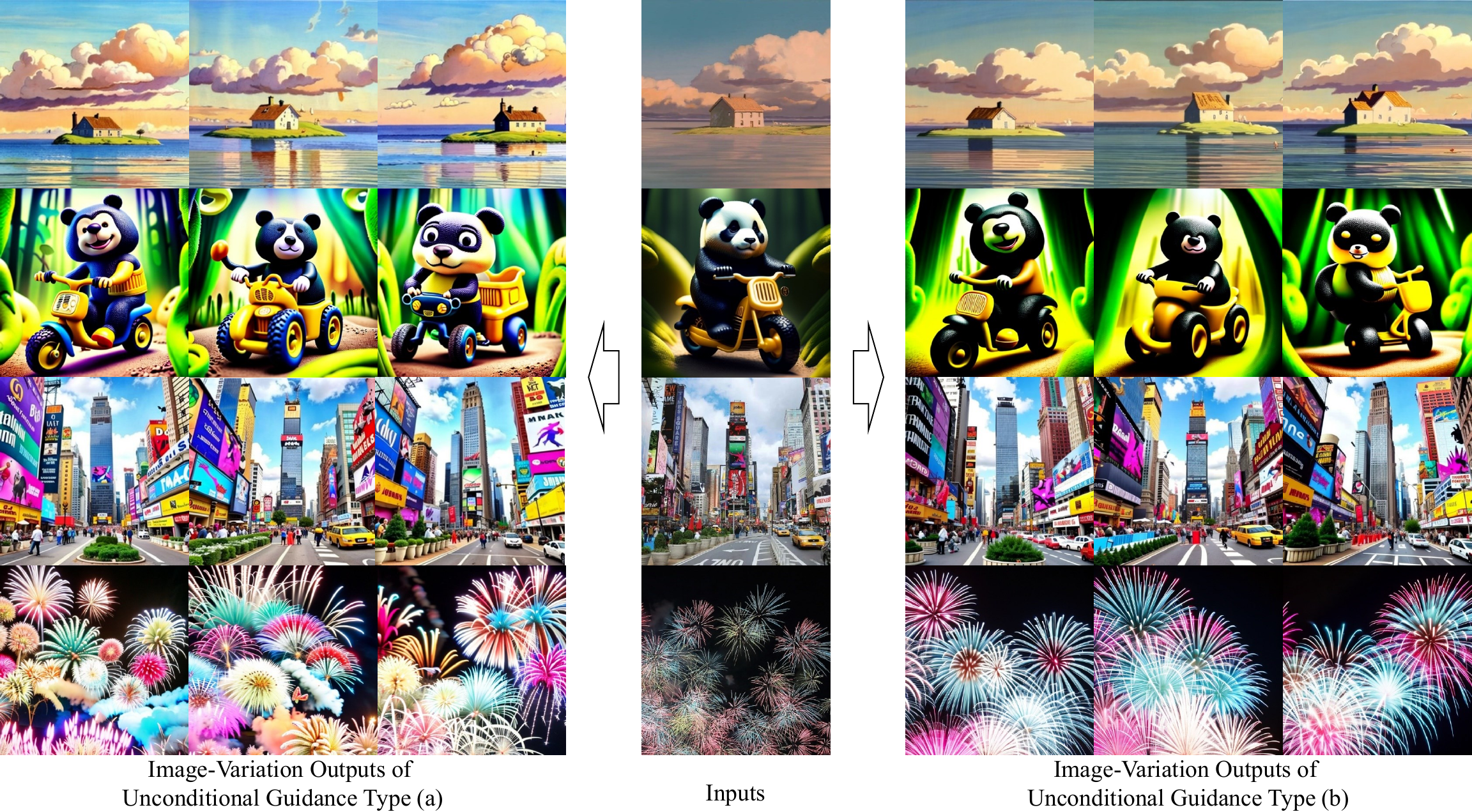}
    \caption{This figure shows the image-variation performance of two types of unconditional guidance described in Appendix~\ref{sec:supp_iv_ug}. The top two rows are cases where type (a) yields better results, while the bottom two rows show that type (b) yields better results. }
    \label{fig:supp_oriug_vs_newug}
\end{figure}

\noindent As shown in Figure~\ref{fig:supp_oriug_vs_newug}, both methods have pros and cons: Option (a) tends to highlight content and style in the reference image, and thus its results are more art-focused with better color contrast. Option (a) may also yield slightly better-performing disentanglement and dual-context because it sensitively captures details from input contexts and ``magnify" them in the output. While this option sometimes may ``over-react'', which results in unbearably color and structure distortions. Conversely, Option (b) is a more robust solution that performs better on photorealistic inputs and generates outputs closer to reference images with fewer distortions.

\subsection{Image-Variation with ControlNet}

\begin{figure}[t]
\centering
    \begin{subfigure}[b]{0.8\textwidth}
        \centering
        \includegraphics[width=\textwidth]{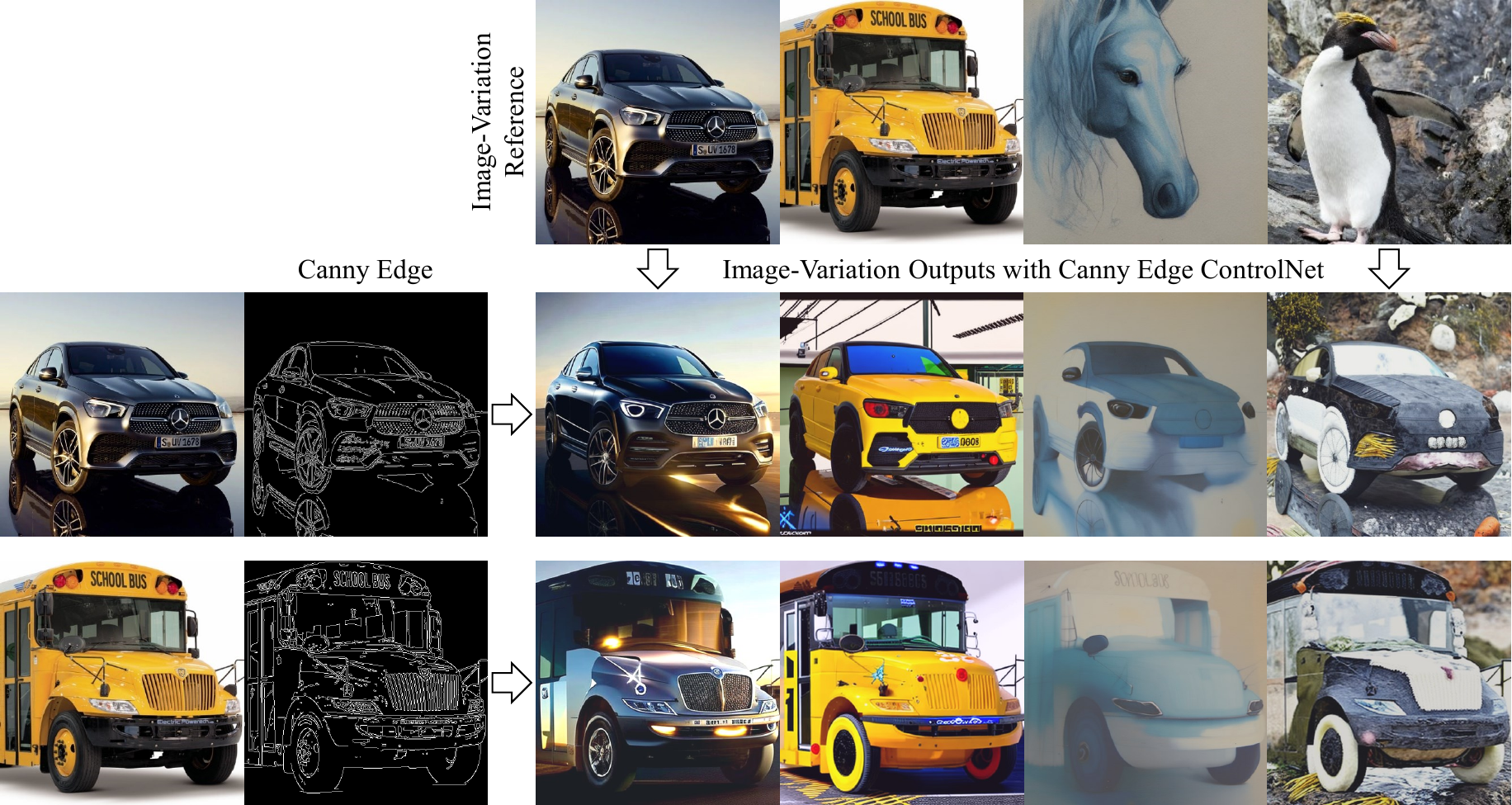}
        \subcaption{VD's Image-variation demo with canny edge ControlNet}
        \vspace{0.3cm}
    \end{subfigure}
    \\
    \begin{subfigure}[b]{0.8\textwidth}
        \centering
        \includegraphics[width=\textwidth]{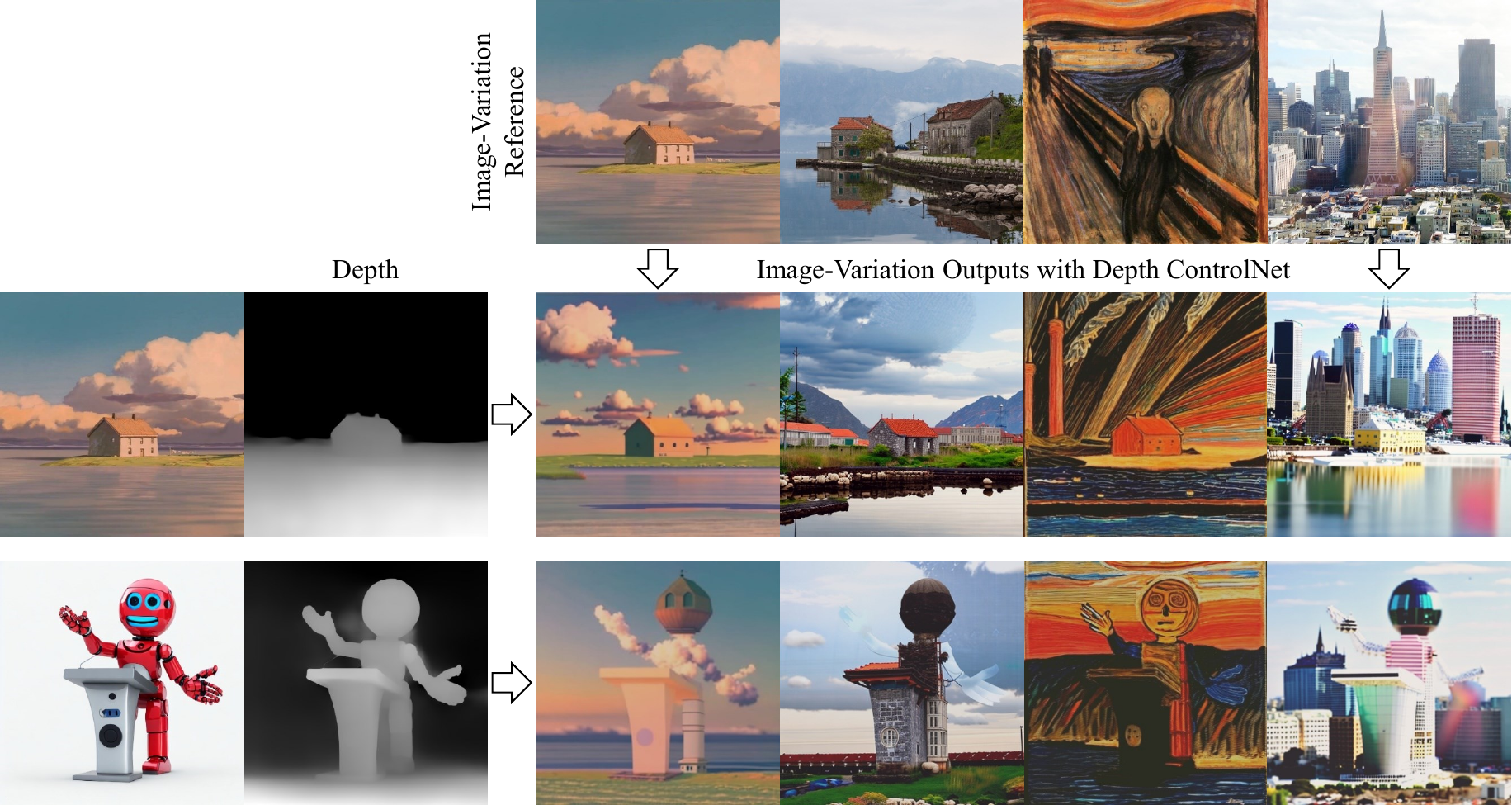}
        \subcaption{VD's Image-variation demo with depth ControlNet}
        \vspace{0.3cm}
    \end{subfigure}
\caption{This figure shows the demo of our VD's image-variation results with ControlNet. A new type of image CLIP, CLIP-PA, is used to generate these results.}
\vspace{-0.1cm}
\label{fig:supp_ivctl}
\end{figure}

ControlNet~\cite{controlnet} and similar techniques such as~\cite{t2i_adapter, alibaba_composer} have recently proposed a practical image editing solution involving pretrained text-to-images models and adaptive networks. One benefit of ControlNet is once an adaptive network (\ie control network) is set up, it can be easily transferred to other text-to-image models without further effort in training. Such convenience inspires us to combine the same adaptive network strategy with VD's image-variation, forming a new application for prompt-free controllable image generation. 

In Figure~\ref{fig:supp_ivctl}, we show the performance of this new application with canny edge and depth ControlNet. The usage of these ControlNet closely follows text-to-image approaches:

\setlist[enumerate,1]{leftmargin=1.5cm, rightmargin=1.5cm}
\begin{enumerate}[label=(\alph*)]
    \item We first prepare a well-trained image variation model under VD's framework.
    \vspace{-0.5em}
    \item We download the pretrained ControlNets and load them together with VD.
    \vspace{-0.5em}
    \item We then control the image-variation process under the guidance of these ControlNets just like in text-to-image. The sole difference is we don't need any prompts.
\end{enumerate}

One thing to notice is that the image-variation model we use for ControlNet slightly alters from the default version of VD, in which we remove the positional embedding layers from the CLIP image encoder to make it a position-agnostic CLIP (CLIP-PA). We then prune VD, letting the image-variation flow, and finetune it with the new CLIP-PA. The finetuned checkpoint is then a position-agnostic image-varation model. Once we have this new model, we can add ControlNet's adaptive network as text-to-image approach. The final outputs will then be a combination of ControlNet's structure hint and VD's semantic and style.

\section{Data and Training}

\subsection{Laion2B Prompt Cleaning}
\label{sec:supp_prompt_clearning}

As mentioned in the main article Section 4.1, we cleaned text prompts from Laion2B in order to include Optimus VAE for the to-text flows in VD. Our rules of cleaning the prompts are the followings:

\setlist[enumerate,1]{leftmargin=1.5cm}
\begin{enumerate}[label=(\alph*)]
    \item Remove all HTTP links, URLs, and email addresses
    \vspace{-0.5em}
    \item Remove HTML syntax.
    \vspace{-0.5em}
    \item Remove unnecessary contents included by square or curly brackets.
    \vspace{-0.5em}
    \item Remove unnecessary symbols such as dashes, slashes, and underscores.
    \vspace{-0.5em}
    \item Remove all kinds of quotes but keep \textit{'s}.
\end{enumerate}

\noindent By cleaning these captions, we were able to train VD's to-text flows in a more robust way. We did not apply such prompt cleaning when training VD on its to-image flows.

\subsection{Alternative Training}

\begin{figure}[t]
\centering
    \begin{subfigure}[b]{0.8\textwidth}
        \centering
        \includegraphics[width=\textwidth]{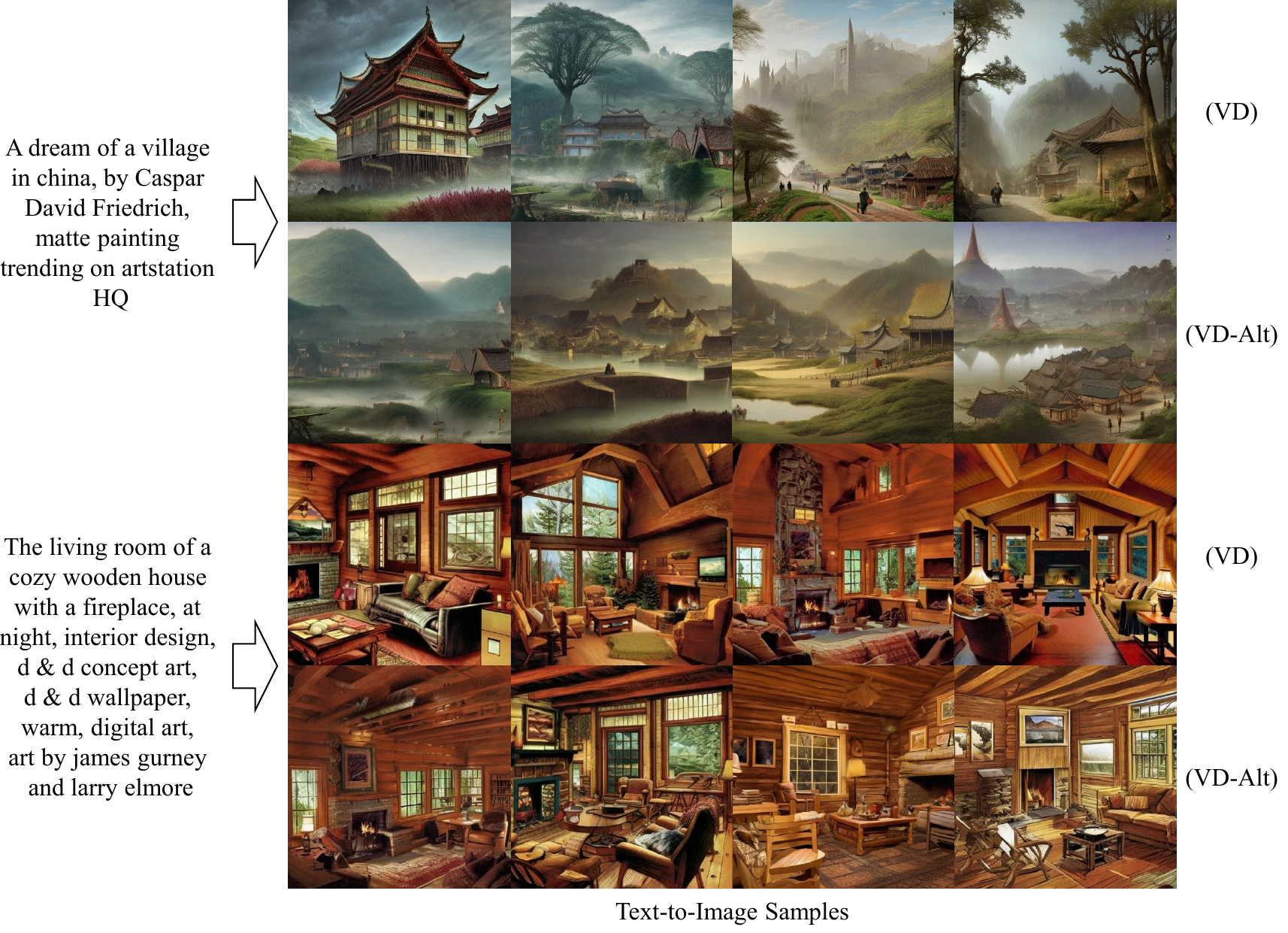}
    \end{subfigure}
    \par\medskip
    \begin{subfigure}[b]{0.8\textwidth}
        \centering
        \includegraphics[width=\textwidth]{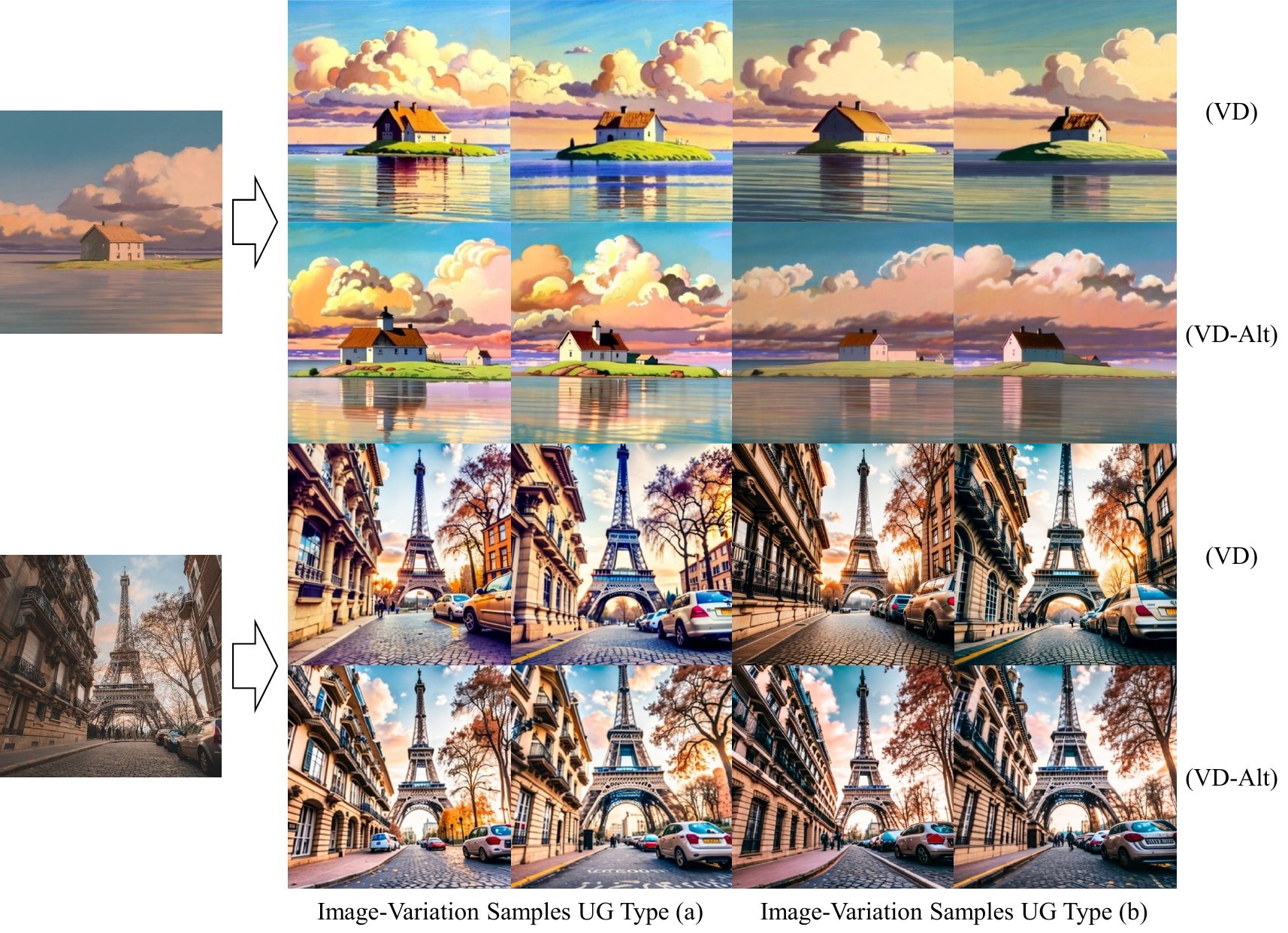}
    \end{subfigure}
\caption{The qualitative comparison between VD and VD-Alt on text-to-image and image-variation. For image-variation, We show both results using unconditional guidance (UG) type (a) and (b) described in Appendix~\ref{sec:supp_iv_ug}.}
\label{fig:supp_vd_alt}
\end{figure}

In Section 4.2, we mentioned that VD's training follows a progressive rule in which we train a single-flow VD, a dual-flow VD, and finally, the four-flow VD in order. Recall that VD's single-flow model is an image-variation model, which means that image-variation is the ``beginning task'' VD first learns. This happened to be the rule we followed in the main paper, but it did not stop other possible ways of training. In fact, training VD can be more flexible. To demonstrate such flexibility, we alternatively trained a VD in which we set text-to-image as the ``beginning task". We compare the final results of this alternative VD model (labeled as VD-Alt) with the paper model (labeled as VD) in Figure~\ref{fig:supp_vd_alt}, in which both yield to similar performance.

\begin{figure}[h]
    \centering
    \includegraphics[width=0.7\textwidth]{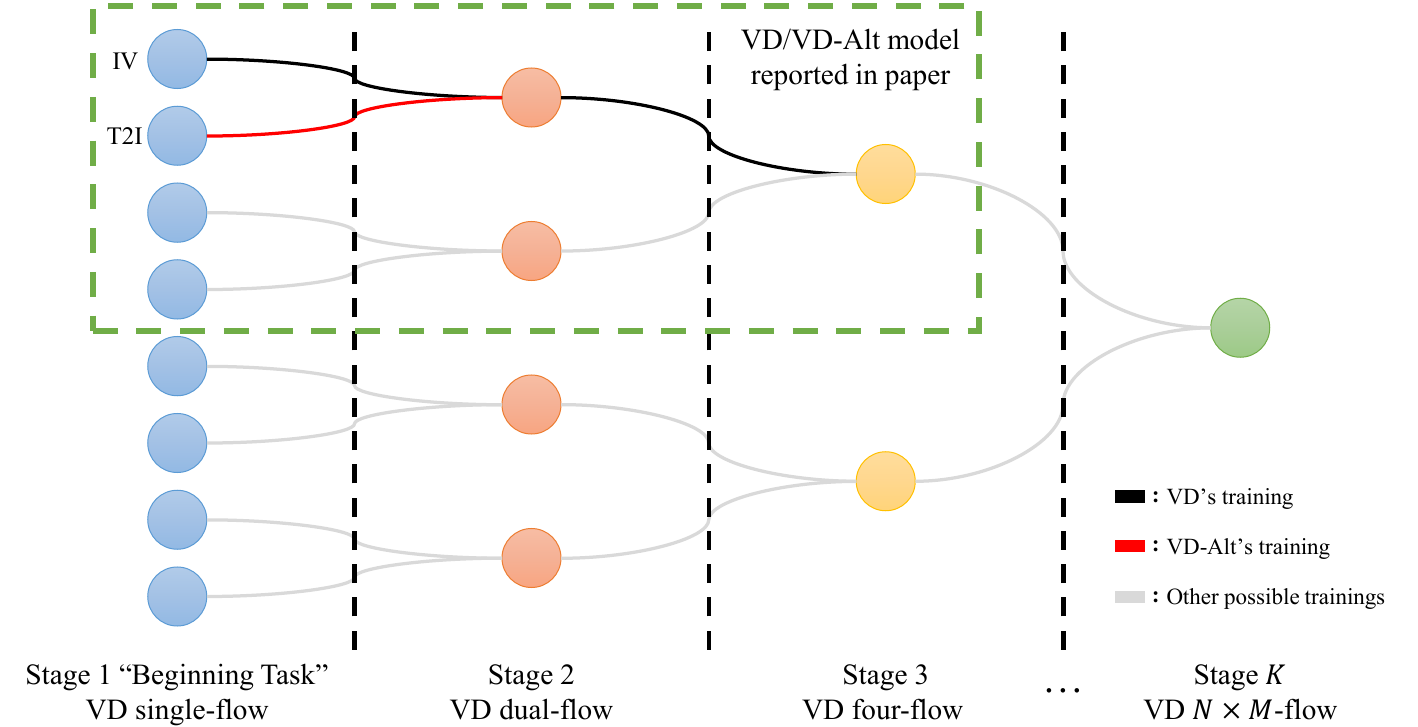}
    \caption{A graphic explanation of possible ways of training the generalized VD with $M$ types of context that supports $N$ types of outputs. The green dash box crop out the current VD/VD-Alt mentioned in the main paper and the appendix. VD's training shows the black paths starting from IV (image-variation), while VD-Alt's training shows an altered red path from T2I (text-to-image). }
    \label{fig:supp_vd_train_order}
\end{figure}

The success of VD-Alt reveals that there may exist many feasible training rules for the multi-flow multimodel diffusion models with $M$ context $N$ outputs. A graphic explanation of the collections of rules is illustrated in Figure~\ref{fig:supp_vd_train_order}. Given that our VD and VD-Alt all yield good performance, we prompt researchers to further explore the flexibility of training VD, which could also be one of the exciting properties of universal AI.

\subsection{Loss Curves}

\begin{figure}[h]
\centering
    \hspace*{\fill}
    \begin{subfigure}[b]{0.4\textwidth}
        \centering
        \resizebox{\columnwidth}{!}{
    \begin{tikzpicture}
    \begin{axis}[
        xlabel=Iterations $\times10^4$,
        grid=both,
        grid style={line width=.1pt, draw=gray!50},
        y tick label style={/pgf/number format/fixed zerofill, /pgf/number format/precision=4},
        ymin=0.12,
        ymax=0.145,
        width=12cm,height=8cm,
    ]
    \addplot [cyan, mark=nomark] table[y=Loss, x=Itern] {data/loss0_0.dat};
    \addlegendentry{$\mathcal{L}_{\text{IV}}$}
    \end{axis}
    \end{tikzpicture}
}
        \vspace{-0.3cm}
        \subcaption{VD single-flow trained on Laion2B Resolution 256}
        \vspace{0.3cm}
    \end{subfigure}
    \hfill
    \begin{subfigure}[b]{0.4\textwidth}
        \centering
        \resizebox{\columnwidth}{!}{
    \begin{tikzpicture}
    \begin{axis}[
        xlabel=Iterations $\times10^3$,
        grid=both,
        grid style={line width=.1pt, draw=gray!50},
        y tick label style={/pgf/number format/fixed zerofill, /pgf/number format/precision=4},
        ymin=0.110,
        ymax=0.145,
        width=12cm,height=8cm,
    ]
    \addplot [cyan, mark=nomark] table [y=Loss, x=Itern]{data/loss0_1.dat};
    \addlegendentry{$\mathcal{L}_{\text{IV}}$}
    \end{axis}
    \end{tikzpicture}
}
        \vspace{-0.3cm}
        \subcaption{VD single-flow trained on Laion2B Resolution 512}
        \vspace{0.3cm}
    \end{subfigure}
    \hspace*{\fill}
    \\    
    \hspace*{\fill}
    \begin{subfigure}[b]{0.4\textwidth}
        \centering
        \resizebox{\columnwidth}{!}{
    \begin{tikzpicture}
    \begin{axis}[
        xlabel=Iterations $\times10^4$,
        grid=both,
        grid style={line width=.1pt, draw=gray!50},
        y tick label style={/pgf/number format/fixed zerofill, /pgf/number format/precision=4},
        ymin=0.12,
        ymax=0.145,
        width=12cm,height=8cm,
    ]
    \addplot [cyan, mark=nomark] table [y=loss_imagectx, x=Itern]{data/loss1_0.dat};
    \addlegendentry{$\mathcal{L}_{\text{IV}}$}
    \addplot [blue, mark=nomark] table [y=loss_textctx, x=Itern]{data/loss1_0.dat};
    \addlegendentry{$\mathcal{L}_{\text{T2I}}$}
    \end{axis}
    \end{tikzpicture}
}
        \vspace{-0.3cm}
        \subcaption{VD dual-flow trained on Laion2B Resolution 256}
        \vspace{0.3cm}
    \end{subfigure}
    \hfill
    \begin{subfigure}[b]{0.4\textwidth}
        \centering
        \resizebox{\columnwidth}{!}{
    \begin{tikzpicture}
    \begin{axis}[
        xlabel=Iterations $\times10^3$,
        grid=both,
        grid style={line width=.1pt, draw=gray!50},
        y tick label style={/pgf/number format/fixed zerofill, /pgf/number format/precision=4},
        ymin=0.110,
        ymax=0.145,
        width=12cm,height=8cm,
    ]
    \addplot [cyan, mark=nomark] table [y=loss_imagectx, x=Itern]{data/loss1_1.dat};
    \addlegendentry{$\mathcal{L}_{\text{IV}}$}
    \addplot [blue, mark=nomark] table [y=loss_textctx, x=Itern]{data/loss1_1.dat};
    \addlegendentry{$\mathcal{L}_{\text{T2I}}$}
    \end{axis}
    \end{tikzpicture}
}
        \vspace{-0.3cm}
        \subcaption{VD dual-flow trained on Laion2B Resolution 512}
        \vspace{0.3cm}
    \end{subfigure}
    \hspace*{\fill}
    \\    
    \hspace*{\fill}
    \begin{subfigure}[b]{0.4\textwidth}
        \centering
        \resizebox{\columnwidth}{!}{
    \begin{tikzpicture}
    \begin{axis}[
        xlabel=Iterations $\times10^4$,
        grid=both,
        grid style={line width=.1pt, draw=gray!50},
        yticklabel style={/pgf/number format/fixed zerofill, /pgf/number format/precision=4},
        ymin=0.0,
        ymax=0.2,
        width=12cm,height=8cm,
    ]
    \addplot [cyan, mark=nomark] table [y=loss_v2i, x=Itern]{data/loss2_0.dat};
    \addlegendentry{$\mathcal{L}_{\text{IV}}$}
    \addplot [blue, mark=nomark] table [y=loss_t2i, x=Itern]{data/loss2_0.dat};
    \addlegendentry{$\mathcal{L}_{\text{T2I}}$}
    \addplot [green, mark=nomark] table [y=loss_v2t, x=Itern]{data/loss2_0.dat};
    \addlegendentry{$\mathcal{L}_{\text{I2T}}$}
    \addplot [brown, mark=nomark] table [y=loss_t2t, x=Itern]{data/loss2_0.dat};
    \addlegendentry{$\mathcal{L}_{\text{TV}}$}
    \end{axis}
    \end{tikzpicture}
}
        \vspace{-0.3cm}
        \subcaption{VD four-flow trained on Laion2B Resolution 256}
        \vspace{0.3cm}
    \end{subfigure}
    \hfill
    \begin{subfigure}[b]{0.4\textwidth}
        \centering
        \resizebox{\columnwidth}{!}{
    \begin{tikzpicture}
    \begin{axis}[
        xlabel=Iterations $\times10^3$,
        grid=both,
        grid style={line width=.1pt, draw=gray!50},
        yticklabel style={/pgf/number format/fixed zerofill, /pgf/number format/precision=4},
        ymin=0.0,
        ymax=0.2,
        width=12cm,height=8cm,
    ]
    \addplot [cyan, mark=nomark] table [y=loss_v2i, x=Itern]{data/loss2_1.dat};
    \addlegendentry{$\mathcal{L}_{\text{IV}}$}
    \addplot [blue, mark=nomark] table [y=loss_t2i, x=Itern]{data/loss2_1.dat};
    \addlegendentry{$\mathcal{L}_{\text{T2I}}$}
    \addplot [green, mark=nomark] table [y=loss_v2t, x=Itern]{data/loss2_1.dat};
    \addlegendentry{$\mathcal{L}_{\text{I2T}}$}
    \addplot [brown, mark=nomark] table [y=loss_t2t, x=Itern]{data/loss2_1.dat};
    \addlegendentry{$\mathcal{L}_{\text{TV}}$}
    \end{axis}
    \end{tikzpicture}
}
        \vspace{-0.3cm}
        \subcaption{VD four-flow trained on COYO Resolution 512}
        \vspace{0.3cm}
    \end{subfigure}
    \hspace*{\fill}
\vspace{-0.3cm}
\caption{This figure shows the loss curves of VD single-flow, dual-flow, and four-flow during their training phases.}
\label{fig:supp_loss}
\end{figure}
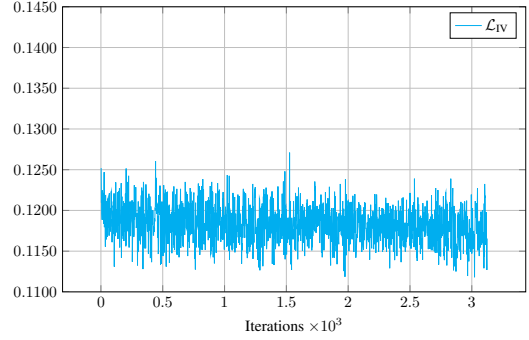
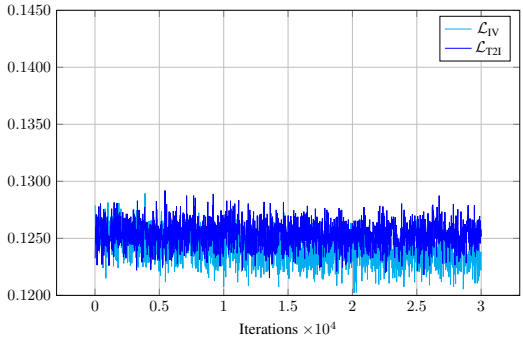
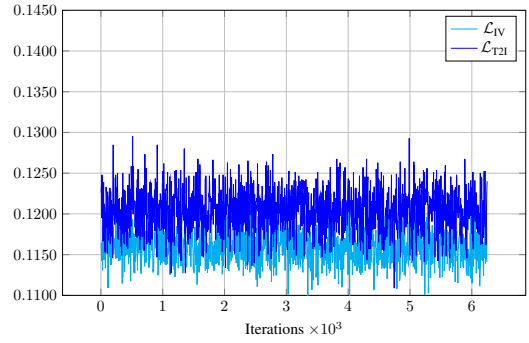
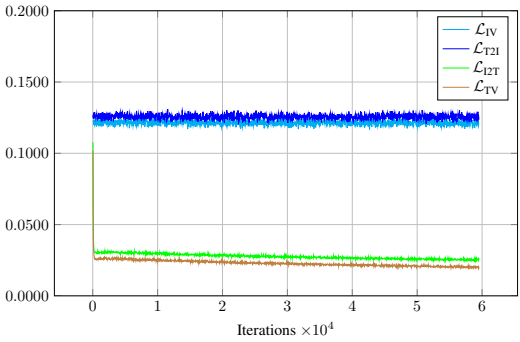
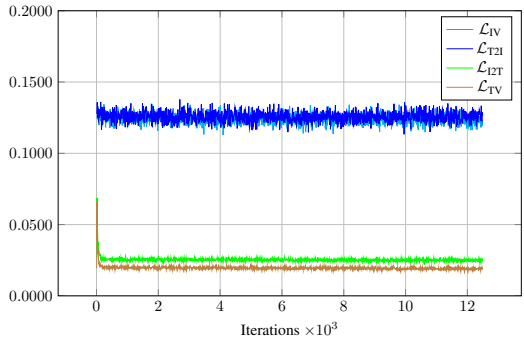

We reveal VD's train-time loss curves in Figure~\ref{fig:supp_loss}. All experiments were carried out using a single node with 8 A100 GPUs (80G memory). To make the effective batch size matches the batch size we mentioned in the main paper Section 4.2 (\ie 2048, 1024, and 512), we utilized the gradient accumulation technique in which we performed multiple backpropagations with one gradient update. The batch per GPU for single backpropagations is 64 for resolution 256 and 16 for resolution 512. The gradient accumulation loop can then be calculated as:
\vspace{0.5cm}

\begin{equation}
    \text{Gradient Accumulation Loop} = \frac{\text{Effective Batch Size}}{\text{Batch per GPU} \times 8} \quad \eg =4 \text{ for single-flow 256 training}
\end{equation}

\section{Text-Variation Performance}

We show the performance of our text-variation results in Table~\ref{table:supp_text_variation},  in which a total number of 8 outputs are sampled from VD for each input. 

\begin{table}[h!]
\centering
\resizebox{\columnwidth}{!}{
    \begin{tabular}{
            L{4cm}
            L{1cm}
            L{5cm}
            L{5cm}
            L{5cm}
            L{5cm}}
        \toprule
            Inputs
            &
            & \multicolumn{4}{c}{Outputs}
            \\
        \midrule
        \multirow{2}{*}{\textbf{\shortstack[l]{blue and yellow balloons in \\the sky}}}
            &
            & two blue balloons and yellow balloons flying in the blue sky and the yellow sky in the sky above the blue sky.
            & two blue balloons and blue balloons flying under the yellow star and the blue and the yellow lights in the sky.
            & blue balloons and the yellow blue balloons and blue sky blue sky behind the sky.
            & two balloons blue and the yellow balloons and blue balloons blue up the yellow sky three above the sky in the sky.
            \\
            \cmidrule(l{0.5em}r{0.5em}){3-6}
            &
            & yellow blue balloons and two blue balloons and the yellow balloons in the sky above blue blue.
            & two blue balloons and the blue balloons in the blue and yellow balloons in the yellow and the green sky.
            & two blue balloons and yellow balloons yellow in the blue sky and the large sky at the sky.
            & two blue balloons and blue lights in the yellow and blue sky blue sky near the sky in the sky.
            \\
        \midrule
        \multirow{2}{*}{\textbf{\shortstack[l]{man standing on the beach \\near sea}}}
            &
            & man standing on the beach standing on the beach on the city beach.
            & man standing on the beach standing at the beach on the beach beach.
            & man standing on the beach standing on the beach standing at the city beach.
            & man standing on the beach standing on the beach on the beach beach.
            \\
            \cmidrule(l{0.5em}r{0.5em}){3-6}
            &
            & man standing on the beach standing on the beach near the beach.
            & man standing on the beach standing on the beach standing on the beach at the beach.
            & man standing on the beach standing on the beach on the beach.
            & man standing on the beach standing on the beach at the beach
            \\
        \midrule
        \multirow{2}{*}{\textbf{grand nebula in the universe}}
            &
            & star giant birds in the huge galaxy circle the planet in the cloud
            & giant star wars in the beautiful moon of the giant symphony in the earth.
            & huge star in the crystal world in the big stadium
            & giant star the great moon in the a huge galaxy in the giant field.
            \\
            \cmidrule(l{0.5em}r{0.5em}){3-6}
            &
            & giant star stars in the huge moon the a huge field in the beautiful city.
            & the giant bird in the giant the huge world in the moon shaped.
            & giant star the moon in the huge galaxy field in the giant field.
            & giant birds the world in the beautiful star planet in the stadium.
            \\
        \midrule
        \multirow{2}{*}{\textbf{\shortstack[l]{realistic scenery of Houston \\texas city view under a starry \\sky in hyperrealistic style and \\ultra HD, 8k}}}
            &
            & dramatic urban landscape of a city construction photo with city marquee and cloudy buildings in the foreground, a bright buildings movie studio imaging the scene in
            & beautiful skyscaw electronic photos star crystal-lit city skyline of a city scene in a scenic highway with a bright city skyline, home cabin on
            & a dark-park high-lit city skyline scene behind huge city laid buildings and landscapes, scoredReport artist dressed in the lovely backdrop,
            & downtown drone shot sharply with a very bright and colorful skyline city background, a backdrop with technical snowy skyline city skyline, the musical play a
            \\
            \cmidrule(l{0.5em}r{0.5em}){3-6}
            &
            & a competitive picture through a bright city cinema camera with a massive neon, big city mountains and city skyline buildings in the
            & a scenic city photo photographs show extreme bright sunshine, a hotel lights exciting city a skyline city at a peak, sky construction big city, with
            & city photography studio Kill a beautiful cityscape shot up close and with an urban theater star skyscrap backdrop with a city skyline square, the
            & scenic blue skysct star buildings with city and city skyline aerial view in the foreground in the background in a sunny bright urban setting overlooking
            \\
        \midrule
        \multirow{2}{*}{\textbf{a pink car}}
            &
            & a pink car a car
            & a pink car
            & a pink car a car
            & a pink car a car
            \\
            \cmidrule(l{0.5em}r{0.5em}){3-6}
            &
            & a pink car
            & a pink car
            & a pink car a pink car
            & a pink car a pink car
            \\
        \midrule
        \multirow{2}{*}{\textbf{\shortstack[l]{a handsome-looking horse \\rider}}}
            &
            & a well-dressed man a smiling horse riding a horse.
            & a well a handsome horse-m a horse smiling. a handsome horse man riding a horse.
            & a very a handsome a horse man dressed a his horse riding his horse
            & a very a tall a horse-ring horse rider a horse riding his short a horse.
            \\
            \cmidrule(l{0.5em}r{0.5em}){3-6}
            &
            & a well-eired a man a smiling horse riding a horse rider.
            & a well a- dressed horse wearing a happy horse riding a horse rider. a horse. a good horse. a horse
            & a well-looking a young horse-hired man riding a horse.
            & a handsome a horse- aired man riding a attractive horse.
            \\
        \bottomrule
    \end{tabular}
}
\vspace{0.3cm}
\caption{
    This table shows the performance of VD on text-variation. 
}
\label{table:supp_text_variation}
\end{table}

\section{Limitation}
So far, we have shown that VD is a powerful model with outstanding capacities. However, VD still has noticeable limitations in some aspects, such as image-to-text, \etc. In this session, we would like to discuss the limitation of our work as well as future research directions for improvements.

\textbf{Limited Latent Space:} During our experiment, we noticed that VD's major weakness is text generation (\ie image-to-text, text-variation, I2T2I). We believe that such weakness can be largely improved if we expand the scope and capacity of the Optimus VAE~\cite{optimus}. Unlike the AutoKL~\cite{ldm} (\ie the image VAE), which transforms images into latent features with dimension $4\times 64\times 64$, and thus keeps the necessary spatial information, Optimus's latent vectors are 768 single dimension vectors generated using Bert~\cite{bert} which may be inadequate for long text sentences. We believe that a better solution should adapt word locations and orders, forming a latent space of sequences, which makes text generation and restoration in later stages easier. In our experiment, we also noticed that VD tended to generate sentences with repeated descriptions, which partially proves our guess that the corresponding VAE is weak in understanding word locations and orders. 

\textbf{Imperfect Data:} Another issue that limited VD's performance was the imperfect text data we used in training. As mentioned in the main article Session 4.1, we formalized web-scraped prompts and captions with extensive engineering effort. We notice that with these cleaned captions, VD's training procedure on text-generation tasks had become more robust and easier to converge. Therefore, an immediate future target for VD research is to prepare a finetuned dataset that helps VD improve its model accuracy. Meanwhile, the pretrained Optimus VAE also had certain data limits. In our experiment, we noticed that Optimus VAE had difficulty reconstructing Laion2B captions. An example is shown below: "\textit{Assorted Cuff Colors Sandals Platform Leather Unique Ladies 1TO9 Cow Camel Ankle Brown EqBF6TT0}" (Laion2B), "\textit{leatherback canvas females posses exotic fruits terme white grass patchions}" (Optimus Reconstruction). We think the issue comes from a domain shift from Optimus's training data (\ie PTB~\cite{dataset_ptb}, SNLI~\cite{dataset_snli}, Yahoo and Yelp corpora~\cite{dataset_yahoo,dataset_yelp}) to VD's training data (\ie Laion2B~\cite{laion}), in which the former contains normal sentences with good grammar, and the latter contains long and descriptive sentences with online freestyle. Therefore, preparing a finetuned text VAE could be the critical next step for future VD research.

\section{Gallery}
\begin{figure}[b]
    \centering
    \includegraphics[width=0.94\textwidth]{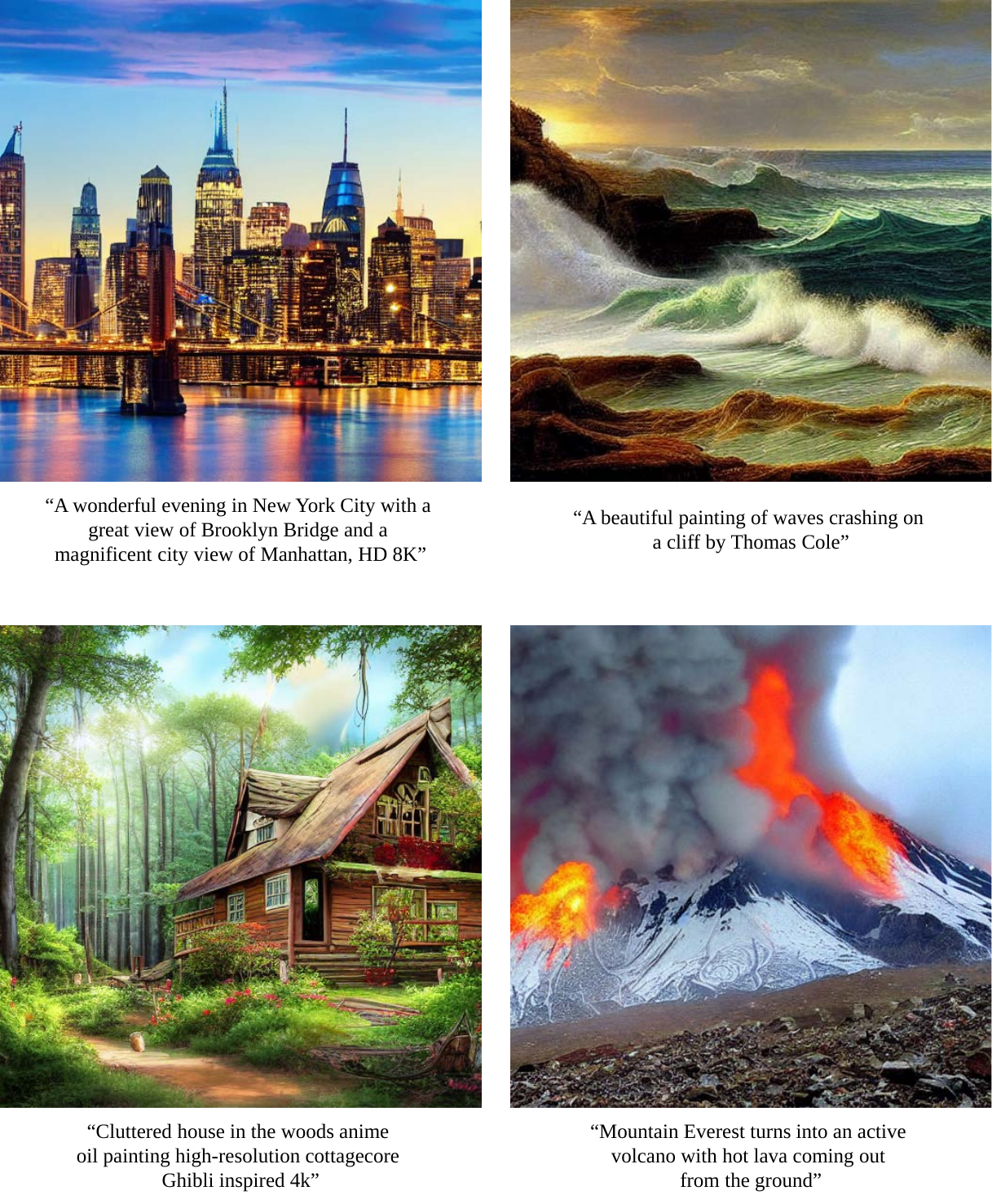}
    \caption{More text-to-image results.}
    \label{fig:supp_gallary_t2i}
\end{figure}

\begin{figure}[t]
    \centering
    \includegraphics[width=0.88\textwidth]{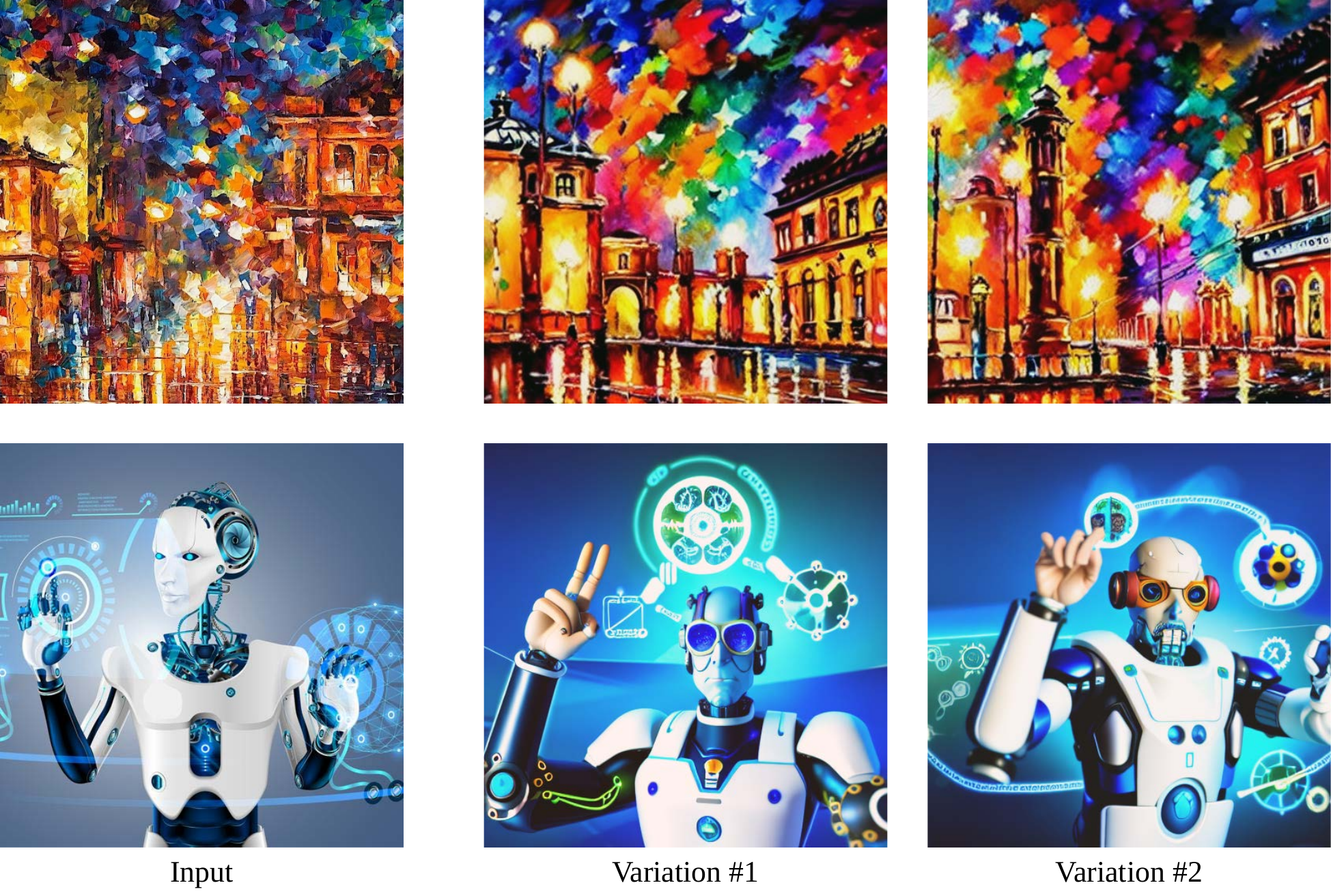}
    \vspace{-0.2cm}
    \caption{More image-variation results.}
    \vspace{-0.2cm}
    \label{fig:supp_gallary_i2i_1}
\end{figure}

\begin{figure}[t]
    \centering
    \includegraphics[width=0.88\textwidth]{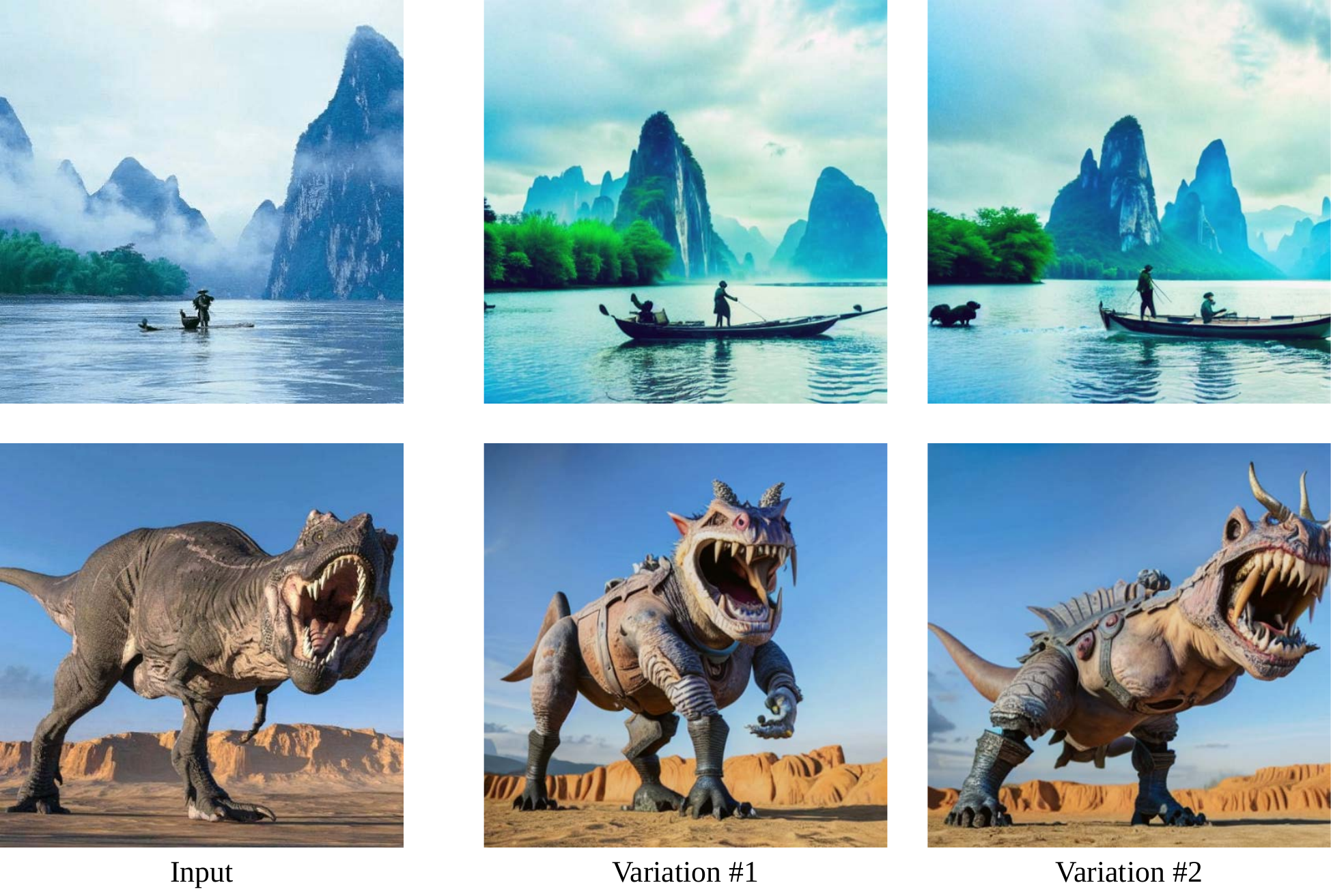}
    \vspace{-0.2cm}
    \caption{More image-variation results with some mild semantic focus to achieve better photorealism.}
    \vspace{-0.2cm}
    \label{fig:supp_gallary_i2i_2}
\end{figure}

\begin{figure}[t]
    \centering
    \includegraphics[width=0.96\textwidth]{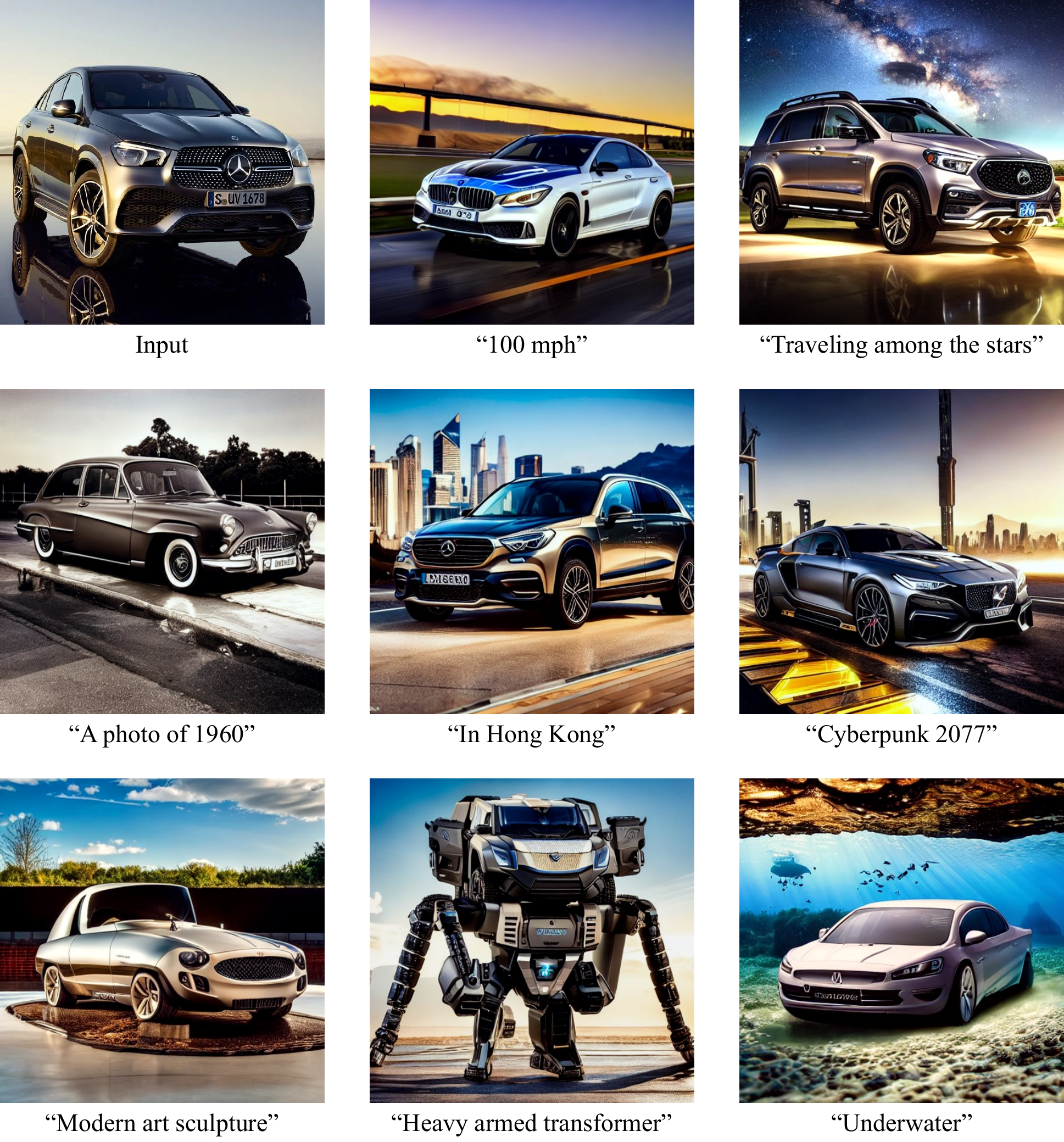}
    \caption{More dual-context blender results, in which the input image is shown at the top left corner, and input prompts are shown as sample labels.}
    \vspace{0.4cm}
    \label{fig:supp_gallary_dg}
\end{figure}

Please see Figure~\ref{fig:supp_gallary_t2i},~\ref{fig:supp_gallary_i2i_1},~\ref{fig:supp_gallary_i2i_2},~\ref{fig:supp_gallary_dg}

\end{appendices}

\end{document}